\documentclass[11pt,a4paper]{article}
\usepackage[hyperref]{acl2019}
\usepackage{times}
\usepackage{latexsym}
\usepackage{mathtools}
\usepackage{amsmath}
\usepackage{amsfonts}
\usepackage{amssymb}
\usepackage{float}
\usepackage{graphicx}
\usepackage{caption}
\usepackage{pgfplots}
\usepackage{svg}
\usepackage[inline]{enumitem}
\pgfplotsset{compat=newest}
\usetikzlibrary{fit,backgrounds,positioning,calc}
\usepackage{url}

\aclfinalcopy 

\everydisplay{\small}

\title{Yahtzee: Reinforcement Learning Techniques for Stochastic Combinatorial Games}

\author{\textbf{Nicholas Pape} \\
  Department of Computer Science \\
  The University of Texas at Austin \\
  \texttt{nickpape@utexas.edu} \\}

\date{2025-12-01}

\begin{document}
\maketitle
\begin{abstract}
    Yahtzee is a classic dice game with a stochastic, combinatorial structure and delayed rewards, making it an interesting mid-scale RL benchmark.
    While an optimal policy for solitaire Yahtzee can be computed using dynamic programming methods, multiplayer is intractable, motivating approximation methods.
    We formulate Yahtzee as a Markov Decision Process (MDP), and train self-play agents using various policy gradient methods:
    REINFORCE, Advantage Actor-Critic (A2C), and Proximal Policy Optimization (PPO), all using a multi-headed network with a shared trunk.
    We ablate feature and action encodings, architecture, return estimators, and entropy regularization to understand their impact on learning.

    Under a fixed training budget, REINFORCE and PPO prove sensitive to hyperparameters and fail to reach near-optimal performance,
    whereas A2C trains robustly across a range of settings. Our agent attains a median score of 241.78 points over 100,000 evaluation games,
    within 5.0\% of the optimal DP score of 254.59, achieving the upper section bonus and Yahtzee at rates of 24.9\% and 34.1\%, respectively.
    All models struggle to learn the upper bonus strategy, overindexing on four-of-a-kind's, highlighting persistent long-horizon credit-assignment
    and exploration challenges.
\end{abstract}

\section{Introduction}
\subsection{Yahtzee as a Reinforcement Learning Benchmark}
While on the surface \textit{Yahtzee} appears to be a trivial dice game \cite{hasbro-2022-yahtzee-rules}, it is actually a complex stochastic optimization problem with combinatorial complexity.

Although there are methods for computing optimal play in \textit{Yahtzee} using dynamic programming, these are computationally expensive and do not scale well to multiplayer settings.
\textit{Yahtzee} offers a rich environment for testing reinforcement learning (RL) solutions due to its combination of a large but manageable state space, randomness, ease of simulation, subtle strategic considerations, and easily identifiable subproblems.
While there have been some efforts to create RL agents for \textit{Yahtzee}, a comprehensive approach using self-play has yet to be published.
It remains an open question of whether deep RL methods can approach optimal performance in full-game \textit{Yahtzee}, and which architectural and training choices most affect learning efficiency and final performance.
Similarly, a robust RL-based solution for multiplayer \textit{Yahtzee} using RL methods has yet to be demonstrated.

\textit{Yahtzee} is an ideal candidate to serve as a bridge between simple toy problems such as \textit{Lunar Lander} \cite{brockman2016openai} and extremely complex games like Go \cite{silver2016alphago}.
Typical small benchmarks often offer low stochasticity and simple combinatorics whereas complex games have intractable state spaces and require massive computational resources and heavy engineering to solve.
\textit{Yahtzee} sits in a middle ground where an analytic optimum exists, but reaching it with RL methods is non-trivial.
These factors make it a challenging yet feasible benchmark for RL research.

\subsection{Objectives}

In this paper we aim to methodically study whether a deep RL agent can achieve near DP-optimal performance in full-game solitaire \textit{Yahtzee} using only self-play,
and how architectural and training choices affect learning efficiency.

Concretely, we ask: \begin{enumerate*}[label=(\roman*)]
    \item How does the trade-off between maximizing single-turn expected score and full-game performance behave?
    \item Can an agent reach optimal performance under a fixed training budget, using only self-play?
    \item Which design choices most affect final performance?
    \item What failure modes exist in learned policies and how could they be addressed?
\end{enumerate*}

\section{Related Work}

\subsection{Policy Gradient Methods and Variance Reduction}
\subsubsection{Return Estimation}
In this paper, we follow notation from \citet{sutton-2018-reinforcement-book} and the policy gradient theorem \cite{sutton-2000-policy-gradient}.

There are multiple methods for assigning credit to actions taken by a policy.
Monte-Carlo (MC) returns $G_{t}^{MC}$ use a summation over the full series of rewards until the end of the episode.
This approach is unbiased but has high variance.
In contrast, TD(0) "Temporal Difference" methods use a "bootstrapped" estimate of future rewards to reduce variance.
Essentially, they only consider received rewards $R$ in a specific time window, and use an estimate from the value function $V(S_{t+1})$ for future rewards beyond that window; this is called the TD estimate \cite{sutton-2018-reinforcement-book}.
This time window can also be adjusted using n-step returns, $G_{t}^{TD(n)}$, which interpolate between MC and TD(0) returns by defining a time horizon $n$ over which to sum rewards before bootstrapping.
A related method is $TD(\lambda)$, which uses an exponentially weighted average of n-step returns, effectively blending multiple time horizons into a single estimate controlled by $\lambda$ \cite{sutton-2018-reinforcement-book}.

While TD estimates are biased (since they rely on future value estimates to be accurate), they have much lower variance than full-episode returns, often improving sample efficiency.
They also provide the benefit of being able to learn online rather than waiting until the end of an episode.

Additionally, pure TD methods can also be viewed as a form of approximate dynamic programming, making them a natural fit for domains where dynamic-programming solutions exist \cite{bertsekas1996neuro}.

\subsubsection{Policy Gradient Methods}
Policy-gradient methods are a family of algorithms which directly optimize a parameterized policy $\pi_{\theta}$ to follow an estimate of the performance gradient.
A simple formulation of this is the REINFORCE algorithm \cite{williams-1992-reinforce}, which uses Monte-Carlo returns $G_{t}^{MC}$ on finite, episodic tasks.
One trick for reducing variance in REINFORCE is to subtract a baseline (often just an average return, but potentially a learned estimate) from an episode's MC return.
This yields an advantage estimate that reduces variance without changing its expectation \cite{weaver2013optimal, greensmith-2004-variance-reduction}.

Actor-critic methods \cite{konda-1999-actorcritic} such as Advantage Actor-Critic (A2C) \cite{mnih-2016-a3c} typically use a TD-style return estimate to update the policy.
These methods learn a separate value function: the critic $V_{\phi}$.
This critic is used directly in the TD return estimate as the bootstrap value estimate for a state.
For these methods, we can define the TD error $\delta_t$ as the difference between the TD estimate and the value estimate for the current state $V(S_t)$.
This $\delta_t$ error is then used as the advantage estimate for a normal policy gradient update \cite{konda-1999-actorcritic}.

Another widely used algorithm, proximal policy optimization (PPO), utilizes a clipped objective $L^{CLIP}(\theta)$
and explicit Kullback-Leibler (KL) divergence control to dramatically reduce variance and ensure stable updates \cite{schulman-2017-ppo}.
PPO uses the Generalized Advantage Estimate (GAE), which is closely related to $TD(\lambda)$, applying a $\lambda$-weighted mixture at the level of advantages \cite{schulman-2016-gae}.

\subsubsection{Other Variance Reduction Techniques}
Aside from return estimation, there is a host of other variance reduction techniques which can be employed for policy gradient methods.

Normalizing advantages across a batch improves gradient conditioning and is common practice \cite{schulman2015trpo}.
Entropy regularization prevents early collapse to suboptimal policies by encouraging exploration via the addition of an explicit entropy bonus term in the loss function \cite{williams-peng-1991-function-optimization, ahmed2019entropy, mnih-2016-a3c, schulman-2017-ppo}.
Gradient clipping is frequently used alongside these techniques to stop rare, but large, gradient updates from destabilizing training \cite{pascanu-2013-rnn-clipping}.
While high variance is unavoidable in deep reinforcement learning, poor performance can often be linked to numerical instability rather than inherent flaws in algorithmic design \cite{bjorck-2022-high-variance};
simple tweaks like normalizing features before activations can dramatically improve stability.

\subsubsection{Reward Shaping}
For games that have sparse, delayed, or hard-to-reach rewards, reward shaping can be used to improve learning speed and stability.
Conceptually, reward shaping involves defining a potential function: $\Phi(s)$.
Environmental rewards are then augmented with the weighted difference in potential between states in a trajectory.
This has been shown to give practitioners the ability to change learning patterns while keeping the underlying optimal policy invariant \cite{ng-1999-reward-shaping}.
The potential function can be hand-designed or learned, although a learned potential function could inadvertently change the optimal policy if not done carefully \cite{devlin-2014-potential-based}.

\subsection{Complex Games}
RL methods have been shown to be successful in games despite high complexity or stochasticity.
In a classic example, \citet{tesauro1995tdgammon} utilized temporal difference learning to achieve superhuman performance in \textit{Backgammon}.
Tetris has also been studied extensively; \citet{bertsekas1996tetris} utilized approximate dynamic programming methods to learn effective policies for the game,
while \citet{gabillon2013tetris} effectively tackled the game using reinforcement learning methods.
\citet{moravcik2017deepstack} demonstrated that \textit{Texas Hold'em} could be effectively learned, despite hidden information.
Many games can be learned well, so long as methods which ensure better exploration are used \cite{osband2016bootstrappeddqn}.
RL methods can also be used to reach high levels of performance on adversarial games, despite their sparse reward structures.
For example, the game of Go, which has a notoriously intractable state space was solved using Monte-Carlo Tree Search and deep value networks \cite{silver2016alphago}.
Subsequent work showed Go could be learned without the use of expert data, purely through self-play \cite{silver2017alphagozero}.
These works establish that RL methods can handle highly stochastic, combinatorial games, suggesting that \textit{Yahtzee} is a natural but underexplored candidate in this family.

\subsection{DP Methods for Yahtzee}
Solitaire \textit{Yahtzee} is a complex game with an upper bound of $~7 \times 10^{15}$ possible states in its state space.
It has a high degree of stochasticity, as dice rolls are the primary driver of state transitions.
Despite this, it has been analytically solved using dynamic programming techniques; \citet{verhoeff-1999-solitaire-yahtzee}, calculated that the average score achieved during ideal play is $254.59$ points.
Later work by \citet{glenn-2006-optimal-yahtzee} optimized the DP approach via symmetries to propose a more efficient algorithm for computing the optimal policy, with a reachable state space of $~5.3 \times 10^8$ states \cite{glenn-2007-solitaire-yahtzee}.

However, adversarial \textit{Yahtzee} remains an open problem.
While \citet{pawlewicz-2011-multiplayer-yahtzee} showed that DP techniques can be expanded to 2-player adversarial \textit{Yahtzee}, they do not scale to more players.
Approximation methods must be utilized for larger player counts. Achieving a near DP optimal score in solitaire \textit{Yahtzee} is a necessary first step towards solving this setting.

\subsection{Reinforcement Learning for Yahtzee}
YAMS attempted to use Q-learning and SARSA to attempt to learn \textit{Yahtzee}, but was not able to surpass $~120$ points median \cite{belaich-2024-yams}.
Likewise, \citet{kang-2018-yahtzee-rl} applied hierarchical MAX-Q, achieving an average score of $129.58$ and a 67\% win-rate over a 1-turn expectimax agent baseline.
\citet{vasseur-2019-strategy-ladders} explored strategy ladders for multiplayer \textit{Yahtzee}, to understand how sensitive Deep-Q networks were to the upper-bonus threshold.
Later, \cite{yuan-2023-two-player-yahtzee} applied Deep-Q networks to the adversarial setting, with moderate success.

Additionally, some recent informal work has reported success using RL methods for \textit{Yahtzee}.
For example, Yahtzotron used heavy supervised pretraining and A2C to achieve an average of $~236$ points \cite{Haefner2021Yahtzotron}.
Although not a true reinforcement learning approach, \citet{DutschkeYahtzee} reports an agent achieving a score of $241.6 \pm 40.7$ after just 8,000 games, using a combination of statistical heuristics.

\section{Problem Formulation}
\subsection{Game Description}
\subsubsection{Rules of Yahtzee}
\textit{Yahtzee} is played with five standard six-sided dice and a shared scorecard containing 13 categories.
Turns are rotated among players. A turn starts with a player rolling all five dice. They may then choose to keep
some dice, re-rolling the remaining ones. This can be repeated two more times, for a total of three total rolls.
After the final roll, the player must select one of the 13 scoring categories to apply to their current dice.
Each category can only be used once and has specific criteria and scoring rules.

\subsubsection{Mathematical Representation of Yahtzee}
\label{sec:yahtzee-definitions}
The space of all possible dice configurations is:
$$\mathcal{D} \in \{1, 2, 3, 4, 5, 6\}^5$$
and the current state of the dice is represented as:
\begin{equation}
    \mathbf{d} \in \mathcal{D}
\end{equation}

In addition, we can represent the score card as a vector of length 13, where each element corresponds to a scoring category:
\begin{equation}
    \mathbf{c} = (c_1, c_2, \ldots, c_{13}) \text{ where } c_i \in \mathcal{D}_i \cup \{\varnothing\}
\end{equation}
where $\varnothing$ indicates an unused category.

Let us also define a dice face counting function which we can use to simplify score calculations:
\begin{align}
    n_v(\mathbf{d})        & = \sum_{i=1}^{5} \mathbb{I}(d_i = v),
    \quad v \in \{1,\dots,6\}                                                 \nonumber \\
    \mathbf{n}(\mathbf{d}) & = \big(n_1(\mathbf{d}),\dots,n_6(\mathbf{d})\big)
\end{align}

Let the potential score for each category be defined as follows (where detailed scoring rules can be found in Appendix~\ref{app:scoring}):
\begin{equation}
    \begin{aligned}
        \mathbf{f}(\mathbf{d}) & =
        \bigl(f_1(\mathbf{d}), f_2(\mathbf{d}), \ldots, f_{13}(\mathbf{d})\bigr)
    \end{aligned}
\end{equation}

The current turn number can be represented as:
\begin{equation}
    t \in \{1, 2, \ldots, 13\}, \quad t = \sum_{i=1}^{13} \mathbb{I}(c_i \neq \varnothing)
\end{equation}

A single turn is composed of an initial dice roll, two optional re-rolls, and a final scoring decision.
Let $r = 0$, with $r \in \{0,1,2\}$ which is the number of rolls taken so far.

Prior to the first roll, the dice are randomized:

$$
    \mathbf{d}_{r=0} \sim U(\mathcal{D})
$$

The player must decide which dice to keep and which to re-roll. Let the player define a keep vector:
\begin{equation}
    \mathbf{k} \in \{0,1\}^5
\end{equation}
where $\mathbf{k}_i = 1$ indicates that die $i$ is kept, otherwise it is re-rolled.

We can then define the transition of the dice state after a re-roll as:
\begin{align*}
    \mathbf{d}' & \sim U(\mathcal{D}),                          \\
    \mathbf{d}_{r+1}
                & = (\mathbf{1} - \mathbf{k}) \odot \mathbf{d}'
    + \mathbf{k} \odot \mathbf{d}
\end{align*}

When $r=2$, the player must choose a scoring category to apply their current dice to. Define a scoring choice mask as a one-hot vector:
\begin{equation}
    \mathbf{s} \in \{0,1\}^{13}, \quad \|\mathbf{s}\|_1 = 1
\end{equation}

For the purposes of calculating the final (or current) score, any field that has not been scored yet can be counted as zero.
We can define a mask vector for this:

\begin{align}
     & \mathbf{u}(\mathbf{c}) \in \{0,1\}^{13}                                                                  \nonumber \\
     & \mathbf{u}(\mathbf{c})_i = \mathbb{I}\bigl(c_i \neq \varnothing\bigr), \quad \forall i = \{ 1, \ldots 13 \}
\end{align}

If a player achieves a total score of 63 or more in the upper section (categories 1-6), they receive a bonus of 35 points:
$$
    B(\mathbf{c}) = \begin{cases}
        35, & \sum_{i=1}^{6} \mathbf{u}(\mathbf{c})_i \cdot \mathbf{c}_i \geq 63 \\
        0,  & \text{otherwise}
    \end{cases}
$$

There is an additional "Joker" bonus rule for multiple Yahtzees, omitted here for brevity.

The player's score can thus be calculated as:
\begin{equation}
    \mathrm{score}(\mathbf{c}) = B(\mathbf{c}) + \big\langle \mathbf{u}(\mathbf{c}), \mathbf{c} \big\rangle
\end{equation}

\subsection{MDP Formulation}
We model \textit{Yahtzee} as a Markov Decision Process $(\mathcal S,\mathcal A,P,R,\gamma)$ \citep{Puterman1994MDP}.

A state is represented as $\mathbf{s} = (\mathbf{d},\mathbf{c},r, t)$, where $\mathbf{d}$ is the current
dice configuration, $\mathbf{c}$ the scorecard, and $r$ the roll index, and $t$ the current turn index
(see Section~\ref{sec:yahtzee-definitions}).

The action is $\mathbf{a} = (\mathbf{k}, \mathbf{s})$, where $\mathbf{k}$ is the keep vector and $s$ is the score category choice.
This can be restated as a parameterization of the policy: $\pi_{\theta}(\mathbf{a}|\mathbf{s}) = \pi_{\theta}(\phi(\mathbf{s}))$,
where $\phi(\mathbf{s})$ is a feature representation of the state $\mathbf{s}$.

The transition function $P$ is is specified in Appendix~\ref{app:transition-function}.

The reward is the change in total score between steps $R_t = \mathrm{score}(c_{t+1}) - \mathrm{score}(c_t)$.

Since we desire to maximize total score at the end of the game, $\gamma = 1$.
\section{Methodology}
\subsection{Tasks}
We optimize two distinct tasks: a single-turn optimization task and a full-game optimization task.
In the full-game optimization task, 13-turn episodes (totalling 39 individual steps) are played to completion.
The objective again is to maximize the total score at the end of the game.
In the single-turn optimization task, the agent is trained to maximize the expected score over a single 3-step turn.
This is a useful subproblem to study, allowing us to iterate on architecture and training choices with shorter training times and without the complications of long-term credit assignment.

\subsection{State Representation \& Input Features}
The design of $\mathbf{\phi}(\mathbf{s}) \rightarrow \mathbf{x}$ is one of the most critical components to the performance of a model \cite{sutton-2018-reinforcement-book}.

Formally, we define the state representation function as
\begin{equation}
    \mathbf{x} = \mathbf{\phi}(\mathbf{s})
\end{equation}
where $\mathbf{s}$ is the raw MDP state (e.g., dice configuration, scorecard, roll index, turn index), and $\mathbf{x}$ is the feature vector or tensor provided as input to the model. The choice of $\mathbf{\phi}$ determines how information from the environment is encoded for learning and inference.
As such, several different representations were tested to evaluate their impact on learning efficiency and final performance.

\subsubsection{Dice Representation}
The dice representation can be encoded in several ways, we attempted 3 different dice representations:

\begin{align*}
     & \phi_{\mathrm{dice}}^{\mathrm{onehot}}(\mathbf{d})   &  & = \bigl[\mathrm{onehot}(d_1), \ldots, \mathrm{onehot}(d_5)\bigr]                                                  \\[4pt]
     & \phi_{\mathrm{dice}}^{\mathrm{bin}}(\mathbf{d})      &  & = \mathbf{n}(\mathbf{d})                                                                                          \\[4pt]
     & \phi_{\mathrm{dice}}^{\mathrm{combined}}(\mathbf{d}) &  & = \bigl[\phi_{\mathrm{dice}}^{\mathrm{onehot}}(\mathbf{d}), \phi_{\mathrm{dice}}^{\mathrm{bin}}(\mathbf{d})\bigr]
\end{align*}

A simple linear representation using the face values of the dice was also tested, but found to perform poorly and was abandoned early in experimentation.

In our experiments, the environment sorts the dice before encoding them, reducing permutation artifacts but potentially introducing rank-based biases.
Permutation-invariant representations are left for future work.

\subsubsection{Scorecard Representation}
There are two important pieces of information $\mathbf{\phi}$ must encode about the scorecard: whether a category is open or closed,
and some form of progress towards the upper bonus \cite{glenn-2007-solitaire-yahtzee}.
$$\phi_{\mathrm{cat}}(\mathbf{c}) = \mathbf{u}(\mathbf{c})$$

We experimented with several ways of encoding the bonus progress, but settled on a simple normalized, clamped sum of the upper section scores:
$$\phi_{\mathrm{bonus}}(\mathbf{c}) = \min\biggl(\frac{1}{63} \sum_{i=1}^{6} c_{i}, 1\biggr)$$

\subsubsection{Computed Features}
There are some key features that can be computed from the raw state, providing these can allow the model to focus on higher-level patterns.

\begin{align*}
    \phi_{\mathrm{progress}}(t)       & = \frac{t}{12}                                                                \\[4pt]
    \phi_{\mathrm{rolls}}(r)          & \in \{0,1\}^3, \quad \|\phi_{\mathrm{rolls}}(r)\|_1 = 1                       \\[4pt]
    \phi_{\mathrm{joker}}(\mathbf{c}) & \in \{0,1\}, \quad \text{(Joker rule active, see Appendix~\ref{app:scoring})}
\end{align*}

We also defined a lock-in feature to indicate whether scoring in a given upper category would secure the upper bonus:
\begin{align*}
    \phi_{\mathrm{lockin}}(\mathbf{d}, \mathbf{c})   & \in \{0,1\}^{6},                                                                                      \\
    \phi_{\mathrm{lockin},k}(\mathbf{d}, \mathbf{c}) & = \mathbb{I}\Bigl\{\sum_{i=1}^{6} \mathbf{u}(\mathbf{c})_i \cdot c_i + f_k(\mathbf{d}) \geq 63\Bigr\}
\end{align*}

\subsection{Action Representation}
\subsubsection{Rolling Action}
\label{sec:rolling-action}
We experiment with two different rolling action representations.
The first is a Bernoulli representation, where each die has an individual binary decision to be re-rolled or held.
The second is a categorical representation, where each of the 32 possible combinations of dice to keep is represented as a unique action.

\[
    a_{\mathrm{roll}} \sim
    \begin{cases}
        \mathrm{Bernoulli}\!\left(\sigma\!\left(f_\theta(\phi(x))\right)\right) \\
        \mathrm{Categorical}\!\left(\mathrm{softmax}\!\left(f_\theta(\phi(x))\right)\right)
    \end{cases}
\]

\subsubsection{Scoring Action}
The scoring action is always a categorical distribution over the 13 scoring categories.
\[
    a_{\mathrm{score}} \sim \mathrm{Categorical}\!\left(\mathrm{softmax}\!\left(f_\theta(\phi(x))\right)\right)
\]

During training, we mask out invalid scoring actions by setting their logits to $-\infty$ before applying the softmax function.
To help with exploration \cite{tijsma2016comparing_exploration}, during training we sample from these distributions; during inference we take the argmax action.

\subsection{Neural Network Architecture}
The neural network uses a unique architecture designed to handle the specific challenges of \textit{Yahtzee}.
The architecture consists of a trunk, followed by heads for the policy and value functions.
We created the network using PyTorch \cite{paszke2019pytorch}, and the training loop is implemented using \texttt{pytorch-lightning} \cite{falcon2019pytorchlightning}.

\subsubsection{Trunk}
The trunk of the network is a standard feedforward architecture with $L$ (typically 2) fully connected hidden layers.
The width of each layer (hidden size $d_h$) is typically 600 neurons, found through empirical hyperparameter tuning (and ablated in Section~\ref{sec:arch-ablations}).
We utilize layer normalization for improved training stability \cite{ba-2016-layernorm, bjorck-2022-high-variance},
dropout with rate $p_d$ for regularization \cite{srivastava2014dropout},
and Swish activations \cite{ramachandran-2017-swish} to introduce stable non-linearities.

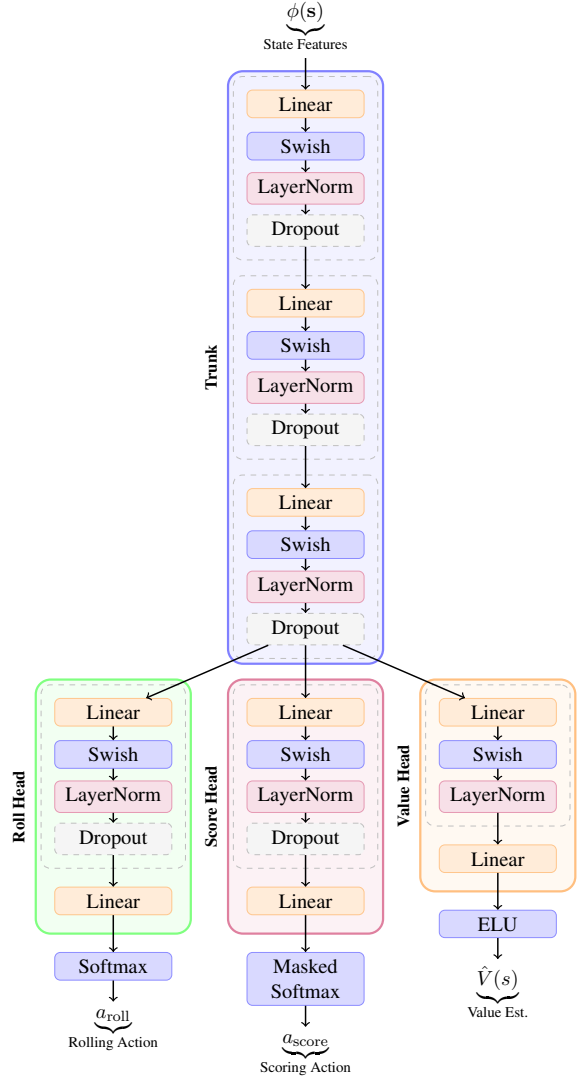
\begin{figure}[tb]
    \centering
    \scalebox{0.70}{
        \begin{tikzpicture}[
                node distance=0.8cm,
                layer/.style={rectangle, draw, minimum width=2.2cm, minimum height=0.5cm, align=center, rounded corners=3pt},
                linear/.style={layer, fill=orange!15, draw=orange!45},
                norm/.style={layer, fill=purple!12, draw=purple!40},
                activation/.style={layer, fill=blue!15, draw=blue!45},
                dropout/.style={layer, fill=gray!8, draw=gray!50, dashed},
            ]

            \node (input) {$\underbrace{\phi(\mathbf{s})}_{\text{State Features}}$};

            \node[linear, below=0.6cm of input] (trunk1_linear) {Linear};
            \node[activation, below of=trunk1_linear] (trunk1_swish) {Swish};
            \node[norm, below of=trunk1_swish] (trunk1_ln) {LayerNorm};
            \node[dropout, below of=trunk1_ln] (trunk1_dropout) {Dropout};

            \node[linear, below=0.8cm of trunk1_dropout] (trunk2_linear) {Linear};
            \node[activation, below of=trunk2_linear] (trunk2_swish) {Swish};
            \node[norm, below of=trunk2_swish] (trunk2_ln) {LayerNorm};
            \node[dropout, below of=trunk2_ln] (trunk2_dropout) {Dropout};

            \node[linear, below=0.8cm of trunk2_dropout] (trunk3_linear) {Linear};
            \node[activation, below of=trunk3_linear] (trunk3_swish) {Swish};
            \node[norm, below of=trunk3_swish] (trunk3_ln) {LayerNorm};
            \node[dropout, below of=trunk3_ln] (trunk3_dropout) {Dropout};

            \node[linear, below left=1.0cm and 1.4cm of trunk3_dropout] (roll_linear) {Linear};
            \node[activation, below of=roll_linear] (roll_swish) {Swish};
            \node[norm, below of=roll_swish] (roll_ln) {LayerNorm};
            \node[dropout, below of=roll_ln] (roll_dropout) {Dropout};

            \node[linear, below=1.0cm of trunk3_dropout] (score_linear) {Linear};
            \node[activation, below of=score_linear] (score_swish) {Swish};
            \node[norm, below of=score_swish] (score_ln) {LayerNorm};
            \node[dropout, below of=score_ln] (score_dropout) {Dropout};

            \node[linear, below right=1.0cm and 1.4cm of trunk3_dropout] (value_linear) {Linear};
            \node[activation, below of=value_linear] (value_swish) {Swish};
            \node[norm, below of=value_swish] (value_ln) {LayerNorm};

            \node[linear, below=0.6cm of roll_dropout] (roll_out_linear) {Linear};
            \node[activation, below=0.7cm of roll_out_linear] (roll_out_softmax) {Softmax};
            \node[below=0.4cm of roll_out_softmax] (roll_out) {$\underbrace{a_{\mathrm{roll}}}_{\text{Rolling Action}}$};

            \node[linear, below=0.6cm of score_dropout] (score_out_linear) {Linear};
            \node[activation, below=0.7cm of score_out_linear] (score_out_softmax) {Masked\\Softmax};
            \node[below=0.4cm of score_out_softmax] (score_out) {$\underbrace{a_{\mathrm{score}}}_{\text{Scoring Action}}$};

            \node[linear, below=0.6cm of value_ln] (value_out_linear) {Linear};
            \node[activation, below=0.7cm of value_out_linear] (value_out_elu) {ELU};
            \node[below=0.4cm of value_out_elu] (value_out) {$\underbrace{\hat{V}(s)}_{\text{Value Est.}}$};

            \begin{scope}[on background layer]
                \node[draw=blue!50, very thick, rectangle, rounded corners=8pt, fit={(trunk1_linear) (trunk1_swish) (trunk1_ln) (trunk1_dropout) (trunk2_linear) (trunk2_swish) (trunk2_ln) (trunk2_dropout) (trunk3_linear) (trunk3_swish) (trunk3_ln) (trunk3_dropout)}, inner sep=0.35cm, fill=blue!5] (trunk_box) {};

                \node[draw=green!50, very thick, rectangle, rounded corners=8pt, fit={(roll_linear) (roll_swish) (roll_ln) (roll_dropout) (roll_out_linear)}, inner sep=0.35cm, fill=green!5] (roll_box) {};

                \node[draw=purple!50, very thick, rectangle, rounded corners=8pt, fit={(score_linear) (score_swish) (score_ln) (score_dropout) (score_out_linear)}, inner sep=0.35cm, fill=purple!5] (score_box) {};

                \node[draw=orange!50, very thick, rectangle, rounded corners=8pt, fit={(value_linear) (value_swish) (value_ln) (value_out_linear)}, inner sep=0.35cm, fill=orange!5] (value_box) {};
            \end{scope}

            \node[draw=gray!50, dashed, rectangle, rounded corners=5pt, fit={(trunk1_linear) (trunk1_swish) (trunk1_ln) (trunk1_dropout)}, inner sep=0.25cm, fill=none] {};

            \node[draw=gray!50, dashed, rectangle, rounded corners=5pt, fit={(trunk2_linear) (trunk2_swish) (trunk2_ln) (trunk2_dropout)}, inner sep=0.25cm, fill=none] {};

            \node[draw=gray!50, dashed, rectangle, rounded corners=5pt, fit={(trunk3_linear) (trunk3_swish) (trunk3_ln) (trunk3_dropout)}, inner sep=0.25cm, fill=none] {};

            \node[draw=gray!50, dashed, rectangle, rounded corners=5pt, fit={(roll_linear) (roll_swish) (roll_ln) (roll_dropout)}, inner sep=0.25cm, fill=none] {};

            \node[draw=gray!50, dashed, rectangle, rounded corners=5pt, fit={(score_linear) (score_swish) (score_ln) (score_dropout)}, inner sep=0.25cm, fill=none] {};

            \node[draw=gray!50, dashed, rectangle, rounded corners=5pt, fit={(value_linear) (value_swish) (value_ln)}, inner sep=0.25cm, fill=none] {};

            \node[left=0.05cm of trunk_box.west, rotate=90, anchor=south, font=\small\bfseries] {Trunk};
            \node[left=0.05cm of roll_box.west, rotate=90, anchor=south, font=\small\bfseries] {Roll Head};
            \node[left=0.05cm of score_box.west, rotate=90, anchor=south, font=\small\bfseries] {Score Head};
            \node[left=0.05cm of value_box.west, rotate=90, anchor=south, font=\small\bfseries] {Value Head};

            \draw[->, thick] (input) -- (trunk1_linear);
            \draw[->, thick] (trunk1_linear) -- (trunk1_swish);
            \draw[->, thick] (trunk1_swish) -- (trunk1_ln);
            \draw[->, thick] (trunk1_ln) -- (trunk1_dropout);
            \draw[->, thick] (trunk1_dropout) -- (trunk2_linear);
            \draw[->, thick] (trunk2_linear) -- (trunk2_swish);
            \draw[->, thick] (trunk2_swish) -- (trunk2_ln);
            \draw[->, thick] (trunk2_ln) -- (trunk2_dropout);
            \draw[->, thick] (trunk2_dropout) -- (trunk3_linear);
            \draw[->, thick] (trunk3_linear) -- (trunk3_swish);
            \draw[->, thick] (trunk3_swish) -- (trunk3_ln);
            \draw[->, thick] (trunk3_ln) -- (trunk3_dropout);

            \draw[->, thick] (trunk3_dropout) -- (roll_linear);
            \draw[->, thick] (trunk3_dropout) -- (score_linear);
            \draw[->, thick] (trunk3_dropout) -- (value_linear);

            \draw[->, thick] (roll_linear) -- (roll_swish);
            \draw[->, thick] (roll_swish) -- (roll_ln);
            \draw[->, thick] (roll_ln) -- (roll_dropout);
            \draw[->, thick] (roll_dropout) -- (roll_out_linear);
            \draw[->, thick] (roll_out_linear) -- (roll_out_softmax);
            \draw[->, thick] (roll_out_softmax) -- (roll_out);

            \draw[->, thick] (score_linear) -- (score_swish);
            \draw[->, thick] (score_swish) -- (score_ln);
            \draw[->, thick] (score_ln) -- (score_dropout);
            \draw[->, thick] (score_dropout) -- (score_out_linear);
            \draw[->, thick] (score_out_linear) -- (score_out_softmax);
            \draw[->, thick] (score_out_softmax) -- (score_out);

            \draw[->, thick] (value_linear) -- (value_swish);
            \draw[->, thick] (value_swish) -- (value_ln);
            \draw[->, thick] (value_ln) -- (value_out_linear);
            \draw[->, thick] (value_out_linear) -- (value_out_elu);
            \draw[->, thick] (value_out_elu) -- (value_out);

        \end{tikzpicture}
    }
    \caption{Overall network architecture with shared trunk and three specialized heads}
    \label{fig:network-architecture}
\end{figure}

\subsubsection{Policy and Value Heads}
We utilize two distinct heads for the rolling and scoring actions, allowing the model to specialize in
each task \cite{tavakoli2018actionbranching, hausknecht2016parameterized}.

We also implement a value head which outputs a scalar baseline for REINFORCE or the value estimate for actor-critic methods.
For the value head, we use a single linear output, constrained with ELU activation to clamp negative value estimates \cite{clevert2016elu}, since negative rewards are not possible in \textit{Yahtzee}.

The rolling and scoring heads implement the distributions from Section~\ref{sec:rolling-action} with a single hidden layer, each.

\subsubsection{Optimization \& Schedules}
We utilize the Adam optimizer \cite{kingma2014adam} with maximum learning rate $\alpha$, typically between $1\times 10^{-4}$ and $1\times 10^{-3}$, tuned empirically.
To improve training stability \cite{liu-2025-warmup-theory, kalra-2024-warmup}, we utilize a warmup schedule over the first 5\% of training,
plateau for 70\% of training, and then linearly decay over the final 25\% of training steps to a minimum ratio $r_{\alpha}$ (typically 5\%) of the maximum \cite{defazio2023optimal, lyle2024normalization}.

\subsubsection{Training Metrics}
To better understand training dynamics, we log several metrics during training.
To monitor the quality of the value network, we log explained variance \cite{schulman2016nutsbolts, schulman-2016-gae}.
To check for policy collapse, we track the policy entropy and KL divergence between policy updates \cite{schulman2016nutsbolts,schulman-2017-ppo},
mask diversity \cite{Hubara2021MaskDiversity},
and the top-k action frequency \cite{sun-etal-2025-curiosity}.
To ensure learning stability, we track gradient norms and clip rate \cite{pascanu-2013-rnn-clipping, Engstrom2020ImplementationMatters}.
Gradient clipping is applied with threshold $\tau_{\mathrm{clip}}$ to prevent destabilizing updates.
To ensure advantages are well-conditioned, we calculate advantage mean and standard deviation \cite{Achiam2018SpinningUp}.
We also monitor standard training metrics such as average reward and loss values.
All metrics were logged to Weights \& Biases \cite{biewald2020wandb}.

\subsection{Reinforcement Learning}
\subsubsection{Reward Shaping}
We also implemented a learned-potential reward shaping mechanism and assess its impact on the model's final performance and ability to learn the bonus.

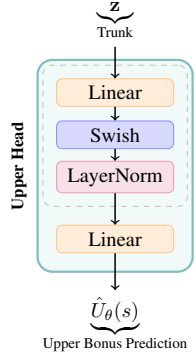
\begin{figure}[tb]
    \centering
    \scalebox{0.70}{
        \begin{tikzpicture}[
                node distance=0.8cm,
                layer/.style={rectangle, draw, minimum width=2.2cm, minimum height=0.5cm, align=center, rounded corners=3pt},
                linear/.style={layer, fill=orange!15, draw=orange!45},
                norm/.style={layer, fill=purple!12, draw=purple!40},
                activation/.style={layer, fill=blue!15, draw=blue!45},
            ]

            \node (input) {$\underbrace{\mathbf{z}}_{\text{Trunk}}$};

            \node[linear, below=0.6cm of input] (upper_linear) {Linear};
            \node[activation, below of=upper_linear] (upper_swish) {Swish};
            \node[norm, below of=upper_swish] (upper_ln) {LayerNorm};

            \node[linear, below=0.6cm of upper_ln] (upper_out_linear) {Linear};
            \node[below=0.7cm of upper_out_linear] (upper_out) {$\underbrace{\hat{U}_\theta(s)}_{\text{Upper Bonus Prediction}}$};

            \begin{scope}[on background layer]
                \node[draw=teal!50, very thick, rectangle, rounded corners=8pt, fit={(upper_linear) (upper_swish) (upper_ln) (upper_out_linear)}, inner sep=0.35cm, fill=teal!5] (upper_box) {};
            \end{scope}

            \node[draw=gray!50, dashed, rectangle, rounded corners=5pt, fit={(upper_linear) (upper_swish) (upper_ln)}, inner sep=0.25cm, fill=none] {};

            \node[left=0.05cm of upper_box.west, rotate=90, anchor=south, font=\small\bfseries] {Upper Head};

            \draw[->, thick] (input) -- (upper_linear);
            \draw[->, thick] (upper_linear) -- (upper_swish);
            \draw[->, thick] (upper_swish) -- (upper_ln);
            \draw[->, thick] (upper_ln) -- (upper_out_linear);
            \draw[->, thick] (upper_out_linear) -- (upper_out);

        \end{tikzpicture}
    }
    \caption{Reward Shaping: Upper bonus prediction head architecture}
    \label{fig:upper-head-architecture}
\end{figure}

First, we implement a new head which predicts the normalized final upper section score at the end of the episode.
This head's architecture is similar to the value head, with a fully connected hidden layer followed by a linear output, with no activation, described in Figure~\ref{fig:upper-head-architecture}.
The target upper score is normalized to the range $[-1, \frac{5}{3}]$ using the formula:

$$
    U_{\mathrm{norm}} = \frac{U_{\mathrm{final}}}{63} - 1 \in \big[-1, \frac{5}{3}\big]
$$

This head is trained using $L_2$ loss:

$$
    \mathcal{L}_{\mathrm{upper}}(\theta) =
    \overbrace{\beta_{\mathrm{regression}}}^{\text{weight}}
    \,
    \biggl\|\hat{U}_\theta(s) - U_{\mathrm{norm}}\biggr\|_2^2
$$

We can convert the normalized score back to a predicted upper score and use it in a potential-based reward shaping function:
\begin{equation}
    \Phi(s) = 35 \cdot \text{clamp}\big(63 \cdot (\hat{U}_\theta(s) + 1), 0, 63\big)
\end{equation}

We then modify the rewards using the potential-based shaping formula \cite{ng-1999-reward-shaping}:
\begin{equation}
    R'(s, a, s') = R(s, a, s') + \beta_{\mathrm{shape}} \cdot (\gamma \Phi(s') - \Phi(s))
\end{equation}

Since the potential function $\Phi$ is changing during training, this may violate Ng's conditions for policy invariance.
However, we wanted to see if it could help the model learn to go for the upper bonus more effectively.

For simplicity, we utilize $r$ to denote the shaped reward $R'$ for the remainder of this paper.

\subsection{Entropy}
\label{sec:entropy}
To encourage exploration, we also add an entropy bonus to the loss function  \cite{williams-peng-1991-function-optimization}. These are held
constant at the start of training then linearly decayed to a final value near the end of training. Different entropy bonuses were used for rolling and scoring actions,
as rolling actions had a tendency to collapse early in training.

Exploration is particularly important for Yahtzee, there are many stable suboptimal policies (e.g., exclusively going for the upper bonus, always going for Yahtzees, etc).
Once the model has figured out how to play the game, it quickly converges without additional exploration incentives.

We can define the entropy bonus as:
\begin{align}
    \mathcal{L}_{\mathrm{entropy}}(\theta)
    = &
    \underbrace{
        \overbrace{\beta_{\mathrm{roll}}}^{\text{weight}}
        \,
        \mathcal{H}\big[\pi_{\theta,\mathrm{roll}}(\cdot \mid s_t)\big]
    }_{\text{rolling action entropy}} \nonumber \\
      & +
    \underbrace{
        \overbrace{\beta_{\mathrm{score}}}^{\text{weight}}
        \,
        \mathcal{H}\big[\pi_{\theta,\mathrm{score}}(\cdot \mid s_t)\big]
    }_{\text{scoring action entropy}}
\end{align}

\subsubsection{Auxilliary Losses}
For all algorithms, we have auxilliary losses for both the shaping head and for entropy:
\begin{align}
    \mathcal{L}_{\mathrm{aux}}(\theta)
    = \mathcal{L}_{\mathrm{upper}}(\theta) + \mathcal{L}_{\mathrm{entropy}}(\theta)
\end{align}

\subsubsection{REINFORCE}
We first implement the REINFORCE algorithm \cite{williams-1992-reinforce} with baseline for single-turn optimization, then attempt to extend it to full-game optimization.
The baseline is the output of the value head, $V_{\phi}(\mathbf{s})$. The loss function is:

\begin{align}
    \mathcal{L}(\theta,\phi)
    = &
    \underbrace{
    \overbrace{- \log \big(\pi_\theta(a_t \mid s_t)\big)}^{\text{negative log likelihood}} \,
    \overbrace{\big(\hat{R}_t - V_\phi(s_t)\big)}^{\text{advantage }}
    }_{\text{policy loss}}                 \nonumber \\
      & +
    \underbrace{
    \overbrace{\lambda_V}^{\text{weight}}
    \|
    V_\phi(s_t) - \hat{R}_t\big
    \|_2
    }_{\text{value loss}}                  \nonumber \\
      & +
    \mathcal{L}_{\mathrm{entropy}(\theta)}
\end{align}

\subsubsection{Advantage Actor-Critic (A2C)}

Second, we utilize an episodic, one-step TD(0) Advantage Actor-Critic (A2C) method. The loss function is:

\begin{align}
    \delta_t
     & =
    \overbrace{r_t}^{\text{reward}}
    +
    \overbrace{\gamma V_\phi(s_{t+1})}^{\text{bootstrap}}
    -
    \overbrace{V_\phi(s_t)}^{\text{current estimate}}
\end{align}
\begin{align}
    \mathcal{L}_{\text{TD-AC}}(\theta,\phi)
    = &
    \underbrace{
    \overbrace{- \log \big(\pi_\theta(a_t \mid s_t)\big)}^{\text{negative log likelihood}}
    \,
    \overbrace{\delta_t}^{\text{TD-error}}
    }_{\text{policy loss}}                 \nonumber \\
      & +
    \underbrace{
    \overbrace{\lambda_V}^{\text{weight}}
    \,
    \big\|
    \delta_t
    \big\|_2^2
    }_{\text{value loss}}                  \nonumber \\
      & +
    \mathcal{L}_{\mathrm{aux}}(\theta)
\end{align}

As this turned out to be the most successfully tuned algorithm, this is the only one for which
we attempted reward shaping.

\subsubsection{PPO}
Lastly, we implement Proximal Policy Optimization (PPO) \cite{schulman-2017-ppo}; we tried this with TD(0) and GAE advantages.
The loss function is:

\begin{align}
    r_t(\theta)
     & =
    \frac{
    \overbrace{\pi_\theta(a_t \mid s_t)}^{\text{current policy}}
    }{
    \underbrace{\pi_{\theta_{\text{old}}}(a_t \mid s_t)}_{\text{behavior policy}}
    }
\end{align}
\begin{align}
    \mathcal{L}(\theta,\phi)
    = &
    \underbrace{
        - \min \bigg\{
        \begin{aligned}
             & r_t(\theta)\,\hat{A}_t,                                                         \\
             & \operatorname{clip}\big(r_t(\theta), 1 - \epsilon, 1 + \epsilon\big)\,\hat{A}_t
        \end{aligned}
        \bigg\}
    }_{\text{policy loss}}                 \nonumber \\
      & +
    \underbrace{
    \overbrace{\lambda_V}^{\text{weight}}
    \big\|
    V_\phi(s_t) - \hat{R}_t
    \big\|_2^2
    }_{\text{value loss}}                  \nonumber \\
      & +
    \mathcal{L}_{\mathrm{entropy}(\theta)}
\end{align}

\subsubsection{Training Regimes}
We analyze several distinct training regimes for \textit{Yahtzee} agents:
\begin{enumerate*}[label=(\roman*)]
    \item REINFORCE directly on the single-turn optimization task and evaluating full-game performance
    \item REINFORCE, TD, and PPO directly on the full-game optimization task
\end{enumerate*}.

During training, we run 1,000 game episodes every 5 epochs (1\% of training) to monitor progress.
These are run using deterministic actions (i.e., taking the action with highest probability) to get a clear picture of the learned policy's performance.
Our final evaluation consists of 100,000 simulated games, providing a robust estimate of the agent's performance.
\section{Results}

\subsection{Single-Turn Results}
\subsubsection{Baseline Single-Turn Performance}
For state representation, the baseline model utilizes:

$$\phi(\mathbf{s}) = \big[\phi_{\mathrm{dice}}^{\mathrm{combined}}(\mathbf{d}), \phi_{\mathrm{cat}}(\mathbf{c}), \phi_{\mathrm{bonus}}(\mathbf{c}), \phi_{\mathrm{rolls}}(r)\big]$$

For outputs, it uses Bernoulli rolling actions and categorical scoring actions.
The single turn model has a short horizon (3 steps); REINFORCE was the natural choice here.
We trained on 260,000 games, \~10 million examples, using a batch size of 1,014 examples, for 10,000 total gradient updates.

\begin{figure}[t]
    \centering
    \begin{tikzpicture}
        \begin{axis}[
                width=0.9\columnwidth,
                height=6cm,
                xlabel={Epochs},
                ylabel={Mean Single-Turn Score},
                xmin=1, xmax=500,
                ymin=3, ymax=20,
                grid=both,
                grid style={dotted},
                tick align=outside,
                tick label style={font=\small},
                label style={font=\small},
                legend style={at={(0.5,1.02)},anchor=south,font=\small,legend columns=2},
                axis y line*=left,
                ylabel style={blue!70!black}
            ]

            \addplot[
                thick,
                blue!70!white
            ] coordinates {
                    (1, 4.159023666381836)
                    (2, 4.631213104724884)
                    (3, 5.262130236625671)
                    (4, 5.828698325157165)
                    (5, 6.294082927703857)
                    (6, 6.7153847217559814)
                    (7, 7.210207152366638)
                    (8, 7.6476332426071165)
                    (9, 7.482988238334656)
                    (10, 7.991863965988159)
                    (20, 11.048668670654298)
                    (30, 11.78254451751709)
                    (40, 11.99215989112854)
                    (50, 12.107544565200806)
                    (60, 12.803254652023316)
                    (70, 12.621745681762695)
                    (80, 12.946893644332885)
                    (90, 12.928846406936646)
                    (100, 13.276035642623901)
                    (110, 13.517455816268921)
                    (120, 13.870266485214234)
                    (130, 13.353550386428832)
                    (140, 13.645710277557374)
                    (150, 13.598225021362305)
                    (160, 13.891568326950074)
                    (170, 13.750591850280761)
                    (180, 13.877219104766846)
                    (190, 14.178994274139404)
                    (200, 13.988905572891236)
                    (210, 14.29792923927307)
                    (220, 13.93964524269104)
                    (230, 14.261390686035156)
                    (240, 14.317751693725587)
                    (250, 14.656509065628052)
                    (260, 14.861982488632203)
                    (270, 14.807988452911378)
                    (280, 15.005917406082153)
                    (290, 14.983432149887085)
                    (300, 14.840236902236938)
                    (310, 15.119378852844239)
                    (320, 14.907692527770996)
                    (330, 14.88816590309143)
                    (340, 15.096449947357177)
                    (350, 15.040236854553223)
                    (360, 15.018935060501098)
                    (370, 14.844083118438721)
                    (380, 15.30458598136902)
                    (390, 15.074852228164673)
                    (400, 14.915384817123414)
                    (410, 14.984763479232788)
                    (420, 14.873225116729737)
                    (430, 15.071154022216797)
                    (440, 15.261834573745727)
                    (450, 15.461982345581054)
                    (460, 15.276035833358765)
                    (470, 15.39393515586853)
                    (480, 15.454881763458252)
                    (490, 15.393343448638916)
                    (500, 15.591420412063599)
                };
            \addlegendentry{Single-Turn Score}

            \addplot[
                dashed,
                blue!70!white,
                thick,
                domain=1:500
            ] {19.59};
            \addlegendentry{DP Optimal}

            \addplot[
                dotted,
                gray!70,
                thick,
                domain=1:500
            ] {3.8};
            \addlegendentry{Random Policy}

            \addlegendimage{thick,red!70!white}
            \addlegendentry{Full-Game Score}

        \end{axis}

        \begin{axis}[
                width=0.9\columnwidth,
                height=6cm,
                ylabel={Mean Full-Game Score},
                xmin=1, xmax=500,
                ymin=39, ymax=260,
                tick align=outside,
                tick label style={font=\small},
                label style={font=\small},
                axis y line*=right,
                axis x line=none,
                ylabel style={red!70!black}
            ]

            \addplot[
                thick,
                red!70!white,
                forget plot
            ] coordinates {
                    (4, 89.93699646)
                    (9, 113.0670013)
                    (14, 134.9360046)
                    (19, 153.6799927)
                    (24, 158.3769989)
                    (29, 157.4949951)
                    (34, 162.6779938)
                    (39, 162.6869965)
                    (44, 168.3280029)
                    (49, 165.9459991)
                    (54, 165.8280029)
                    (59, 171.1269989)
                    (64, 169.3150024)
                    (69, 173.3609924)
                    (74, 175.0670013)
                    (79, 174.5850067)
                    (84, 173.7129974)
                    (89, 174.9400024)
                    (94, 179.522995)
                    (99, 179.2380066)
                    (104, 180.8220062)
                    (109, 180.496994)
                    (114, 179.1670074)
                    (119, 183.1909943)
                    (124, 181.9689941)
                    (129, 178.5579987)
                    (134, 182.2749939)
                    (139, 178.0599976)
                    (144, 181.1329956)
                    (149, 179.8670044)
                    (154, 181.9250031)
                    (159, 182.7960052)
                    (164, 182.3899994)
                    (169, 182.2640076)
                    (174, 184.8040009)
                    (179, 185.6869965)
                    (184, 183.3809967)
                    (189, 184.5890045)
                    (194, 188.1130066)
                    (199, 184.3699951)
                    (204, 183.7799988)
                    (209, 187.7700043)
                    (214, 187.3439941)
                    (219, 184.8800049)
                    (224, 185.878006)
                    (229, 186.2769928)
                    (234, 188.4429932)
                    (239, 192.2519989)
                    (244, 192.6029968)
                    (249, 192.8509979)
                    (254, 193.6679993)
                    (259, 194.1340027)
                    (264, 191.4689941)
                    (269, 194.1349945)
                    (274, 193.3059998)
                    (279, 195.8390045)
                    (284, 194.0599976)
                    (289, 196.3509979)
                    (294, 198.201004)
                    (299, 194.0650024)
                    (304, 195.2200012)
                    (309, 199.1909943)
                    (314, 198.1199951)
                    (319, 196.7449951)
                    (324, 196.6300049)
                    (329, 195.1990051)
                    (334, 193.598999)
                    (339, 198.5839996)
                    (344, 197.1459961)
                    (349, 199.5839996)
                    (354, 195.5440063)
                    (359, 199.9980011)
                    (364, 196.8760071)
                    (369, 193.048996)
                    (374, 198.1450043)
                    (379, 196.0639954)
                    (384, 198.1000061)
                    (389, 198.4129944)
                    (394, 196.6179962)
                    (399, 198.0950012)
                    (404, 198.348999)
                    (409, 196.848999)
                    (414, 198.2680054)
                    (419, 196.6640015)
                    (424, 195.9869995)
                    (429, 199.3150024)
                    (434, 200.3540039)
                    (439, 198.6880035)
                    (444, 199.8220062)
                    (449, 199.8170013)
                    (454, 202.4900055)
                    (459, 198.5119934)
                    (464, 199.8170013)
                    (469, 200.3809967)
                    (474, 200.8209991)
                    (479, 200.0529938)
                    (484, 202.1959991)
                    (489, 200.348999)
                    (494, 202.1139984)
                    (499, 203.1260071)
                };

        \end{axis}
    \end{tikzpicture}
    \caption{Single-turn and full-game performance during training.}
    \label{fig:single-turn-dual-axis}
\end{figure}
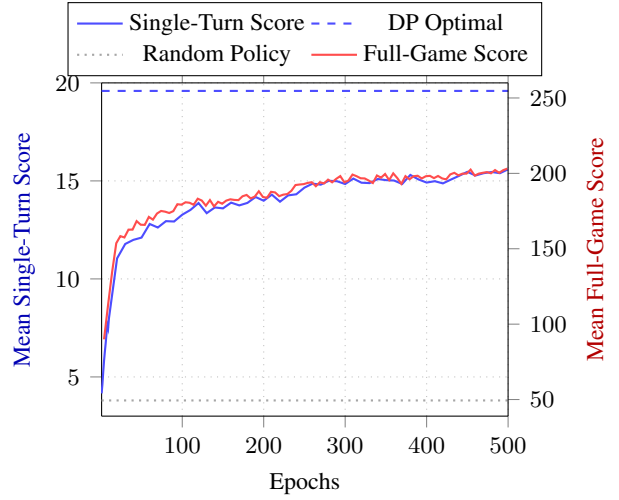
\input{figures/pareto_frontier}

As shown in Figure~\ref{fig:single-turn-dual-axis}, although the single-turn agent does not nearly reach optimal single-turn performance, it performs surprisingly well over the full game; this is likely due to the high correlation between single-turn and full-game optimal actions.
However, we suspected target leakage (selecting parameters and architectures based on full-game performance) could also play a role. This is analyzed in Section~\ref{sec:tradeoff-curve}.

\subsubsection{Single vs Full-game Tradeoff Curve}
\label{sec:tradeoff-curve}

To understand the tradeoff between single-turn and full-game performance, we ablated our model using small changes to various hyperparameters and captured
the resulting performance on both the primary single-turn score, as well as the auxiliary full-game score.

As we suspected, there is a Pareto frontier between these two objectives, as illustrated in Figure~\ref{fig:pareto-frontier}. We can see that full game performance generally increases linearly with single-turn performance.
However, at very high levels of full-game performance, single-turn performance begins to plateau, and even decline slightly.
Since the single-turn model does not have access to the full game context, these are imperfectly optimizing their target objective.
This indicates that selecting hyperparameters for a single-turn model based on full-game performance could indeed be a form of target leakage.

\subsection{Full-Game Results}
\subsubsection{Algorithm Comparison: REINFORCE, A2C, PPO}

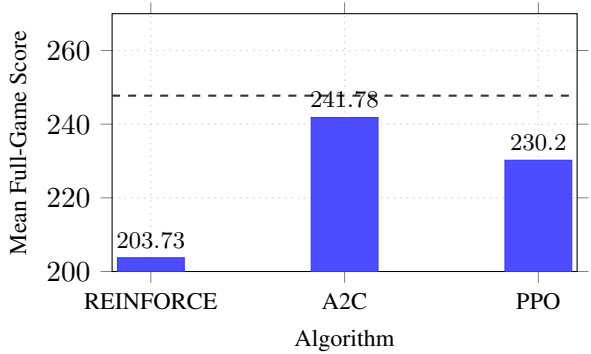
\begin{figure}[t]
    \centering
    \begin{tikzpicture}
        \begin{axis}[
                ybar,
                width=\columnwidth,
                height=5cm,
                xlabel={Algorithm},
                ylabel={Mean Full-Game Score},
                title={},
                symbolic x coords={REINFORCE, A2C, PPO, DP Optimal},
                xtick=data,
                xticklabel style={font=\small},
                ylabel style={font=\small},
                xlabel style={font=\small},
                title style={font=\small},
                bar width=25pt,
                ymin=200, ymax=270,
                grid=both,
                grid style={dotted},
                tick align=outside,
                nodes near coords,
                nodes near coords style={font=\footnotesize, anchor=south},
            ]

            \addplot[
                fill=blue!70!white,
                draw=blue!80,
                error bars/.cd,
                y dir=both,
                y explicit
            ] coordinates {
                    (REINFORCE, 203.73)
                    (A2C, 241.78)
                    (PPO, 230.20)
                };

            \draw[dashed, black!80, thick] (rel axis cs:0,0.682) -- (rel axis cs:1,0.682);

        \end{axis}
    \end{tikzpicture}
    \caption{Final Performance Comparison after training on 1 million games}
    \label{fig:final-performance-comparison}
\end{figure}

\input{figures/baselines_training_graph}

During development, we compared algorithms using a fixed training budget of 250,000 full games played.
Later, we attempt a longer, 1 million full-game training run for our best algorithm.

REINFORCE proved challenging to optimize to high performance levels given our fixed training budget.
It was sensitive to hyperparameters such as the critic coefficient, the entropy bonus, and batch size.
We also found that REINFORCE required more games to converge.
After optimization we were able to achieve reasonable performance; the million game training run scored a mean of 203.7 points.

Our most successful algorithm was TD(0)-style Actor-Critic (A2C). We found is easiest to tune and with an immediate performance boost over REINFORCE.
This was the algorithm we use for the ablation studies. With a training budget of 1 million full-games,
A2C was able to approach DP-optimal performance: scoring 241.8 points average.

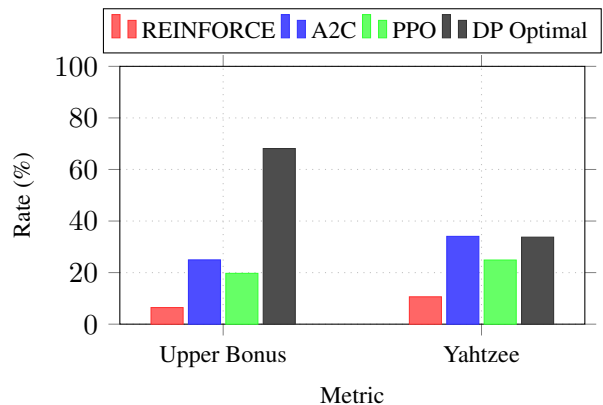
\begin{figure}[b]
    \centering
    \begin{tikzpicture}
        \begin{axis}[
                ybar,
                width=\columnwidth,
                height=5cm,
                xlabel={Metric},
                ylabel={Rate (\%)},
                title={Bonus and Yahtzee Rates by Algorithm},
                symbolic x coords={Upper Bonus, Yahtzee},
                xtick=data,
                xticklabel style={font=\small},
                ylabel style={font=\small},
                xlabel style={font=\small},
                title style={font=\small},
                ymin=0, ymax=100,
                bar width=12pt,
                grid=both,
                grid style={dotted},
                tick align=outside,
                legend style={at={(0.5,1.05)},anchor=south,legend columns=4,font=\small},
                enlarge x limits=0.4,
            ]

            \addplot[
                fill=red!60!white,
                draw=red!80!white
            ] coordinates {
                    (Upper Bonus, 6.44)
                    (Yahtzee, 10.61)
                };
            \addlegendentry{REINFORCE}

            \addplot[
                fill=blue!70!white,
                draw=blue!80!white
            ] coordinates {
                    (Upper Bonus, 24.93)
                    (Yahtzee, 34.05)
                };
            \addlegendentry{A2C}

            \addplot[
                fill=green!60!white,
                draw=green!80!white
            ] coordinates {
                    (Upper Bonus, 19.71)
                    (Yahtzee, 24.87)
                };
            \addlegendentry{PPO}

            \addplot[
                fill=black!70,
                draw=black!80
            ] coordinates {
                    (Upper Bonus, 68.12)
                    (Yahtzee, 33.74)
                };
            \addlegendentry{DP Optimal}

        \end{axis}
    \end{tikzpicture}
    \caption{Bonus and Yahtzee achievement rates for best models of each algorithm}
    \label{fig:bonus-yahtzee-rates}
\end{figure}

We also attempted to use Proximal Policy Optimization (PPO) with TD(0), but found it difficult to tune. Each PPO rollout requires
$k$ epochs of minibatch updates, which significantly increases training time compared to A2C and REINFORCE.
For fair comparison to the other algorithms, reduced the total number of games seen during training by a factor of $k$.
PPO was able to outperform REINFORCE, but was not able to reach A2C performance within our training budget.
It earned a mean score of 230.2 points.
However, it is possible PPO could reach or surpass A2C performance with more extensive hyperparameter tuning.

Figure~\ref{fig:full-game-learning-curves} shows the learning curves for all three algorithms during training.
The final performance comparison is summarized in Figure~\ref{fig:final-performance-comparison}, with detailed bonus and Yahtzee achievement rates shown in Figure~\ref{fig:bonus-yahtzee-rates}.

\subsubsection{Representational Ablations}

\input{figures/dice_representation_bonus}
\begin{figure}[b]
    \centering
    \begin{tikzpicture}
        \begin{axis}[
                width=\columnwidth,
                height=5cm,
                xlabel={Epochs},
                ylabel={Upper Bonus Rate (\%)},
                xmin=0, xmax=500,
                ymin=0, ymax=14,
                grid=both,
                grid style={dotted},
                tick align=outside,
                tick label style={font=\small},
                label style={font=\small},
                legend style={at={(0.5,1.02)},anchor=south,font=\small,legend columns=2}
            ]

            \addplot[
                thick,
                blue!70!white
            ] coordinates {
                    (4, 0) (9, 0) (14, 0.015) (19, 0.057750002) (24, 0.049087502) (29, 0.041724376) (34, 0.05046572) (39, 0.117895862) (44, 0.145211485) (49, 0.138429762) (54, 0.1626653) (59, 0.243265512) (64, 0.236775686) (69, 0.276259333) (74, 0.294820434) (79, 0.29559737) (84, 0.371257767) (89, 0.390569102) (94, 0.391983737) (99, 0.483186177) (104, 0.48570825) (109, 0.502852016) (114, 0.517424217) (119, 0.724810581) (124, 0.871089001) (129, 1.130425672) (134, 1.155861832) (139, 1.117482554) (144, 1.264860157) (149, 1.420131126) (154, 1.492111453) (159, 1.62829475) (164, 1.819050516) (169, 1.771192938) (174, 1.745514001) (179, 1.978686894) (184, 1.801883862) (189, 1.756601282) (194, 1.79311109) (199, 1.704144434) (204, 1.913522754) (209, 1.926494341) (214, 1.982520183) (219, 2.19514217) (224, 2.255870866) (229, 2.367490236) (234, 2.447366679) (239, 2.65026167) (244, 2.687722398) (249, 2.659564038) (254, 2.710629433) (259, 2.709035025) (264, 2.812679785) (269, 3.080777803) (274, 3.338661161) (279, 3.317861994) (284, 3.435182681) (289, 3.564905307) (294, 3.555169511) (299, 3.756894099) (304, 4.258360041) (309, 4.429606049) (314, 4.695165113) (319, 5.190890346) (324, 5.74725688) (329, 6.055168377) (334, 6.241893077) (339, 6.430609187) (344, 6.636017838) (349, 6.900615105) (354, 7.095522811) (359, 7.741194332) (364, 8.290015125) (369, 8.231512942) (374, 8.211786058) (379, 8.405018149) (384, 8.899265398) (389, 9.199375531) (394, 9.079469144) (399, 9.00754883) (404, 8.811416477) (409, 8.614704077) (414, 9.077498437) (419, 9.245873643) (424, 9.268992539) (429, 9.918643715) (434, 9.885847272) (439, 10.05297018) (444, 10.22502477) (449, 10.34127105) (454, 10.5900804) (459, 10.96656825) (464, 11.30158298) (469, 11.28634565) (474, 11.1683938) (479, 11.29313473) (484, 11.44416455) (489, 11.66253981) (494, 11.3531589) (499, 11.57018509)
                };
            \addlegendentry{All Features (Baseline)}

            \addplot[
                thick,
                red!60!white
            ] coordinates {
                    (4, 0) (9, 0) (14, 0) (19, 0) (24, 0) (29, 0.03) (34, 0.040500001) (39, 0.064425001) (44, 0.099761253) (49, 0.084797065) (54, 0.072077505) (59, 0.061265879) (64, 0.067075998) (69, 0.057014598) (74, 0.048462408) (79, 0.071193047) (84, 0.06051409) (89, 0.096436979) (94, 0.081971432) (99, 0.099675717) (104, 0.129724362) (109, 0.170265708) (114, 0.204725853) (119, 0.204016975) (124, 0.218414431) (129, 0.215652267) (134, 0.258304427) (139, 0.234558763) (144, 0.24437495) (149, 0.25271871) (154, 0.259810905) (159, 0.325839276) (164, 0.381963392) (169, 0.414668887) (174, 0.367468554) (179, 0.327348271) (184, 0.323246032) (189, 0.334759128) (194, 0.359545259) (199, 0.335613471) (204, 0.330271452) (209, 0.400730736) (214, 0.415621125) (219, 0.458277964) (224, 0.41953627) (229, 0.401605831) (234, 0.461364958) (239, 0.437160216) (244, 0.401586184) (249, 0.386348258) (254, 0.493396023) (259, 0.464386622) (264, 0.424728629) (269, 0.376019335) (274, 0.379616435) (279, 0.367673972) (284, 0.327522876) (289, 0.443394448) (294, 0.451885281) (299, 0.459102489) (304, 0.435237117) (309, 0.414951552) (314, 0.397708821) (319, 0.398052498) (324, 0.428344627) (329, 0.514092933) (334, 0.556978995) (339, 0.563432149) (344, 0.538917328) (349, 0.548079732) (354, 0.615867772) (359, 0.643487608) (364, 0.681964464) (369, 0.669669798) (374, 0.674219335) (379, 0.723086435) (384, 0.854623473) (389, 0.876429952) (394, 0.834965463) (399, 0.889720651) (404, 0.876262555) (409, 0.909823175) (414, 0.983349713) (419, 0.985847256) (424, 1.017970175) (429, 1.000274645) (434, 1.015233452) (439, 1.102948438) (444, 1.147506186) (449, 1.170380269) (454, 1.249823236) (459, 1.317349758) (464, 1.494747294) (469, 1.585535186) (474, 1.617704901) (479, 1.525049166) (484, 1.596291791) (489, 1.626848015) (494, 1.682820813) (499, 1.730397691)
                };
            \addlegendentry{No Lock-in}

            \addplot[
                thick,
                green!60!black
            ] coordinates {
                    (4, 0) (9, 0.015) (14, 0.01275) (19, 0.0258375) (24, 0.021961875) (29, 0.048667594) (34, 0.056367456) (39, 0.077912338) (44, 0.066225487) (49, 0.056291664) (54, 0.062847915) (59, 0.053420727) (64, 0.075407619) (69, 0.064096476) (74, 0.159482012) (79, 0.225559714) (84, 0.28172576) (89, 0.374466892) (94, 0.393296859) (99, 0.379302332) (104, 0.382406983) (109, 0.445045937) (114, 0.558289054) (119, 0.609545692) (124, 0.63811384) (129, 0.707396768) (134, 0.841287256) (139, 0.835094169) (144, 0.859830044) (149, 0.850855539) (154, 0.813227212) (159, 0.91624313) (164, 1.063806657) (169, 1.144235662) (174, 1.212600316) (179, 1.435710276) (184, 1.610353756) (189, 1.638800686) (194, 1.692980583) (199, 1.844033502) (204, 2.13742847) (209, 2.431814185) (214, 2.367042057) (219, 2.416985756) (224, 2.399437885) (229, 2.264522203) (234, 2.269843865) (239, 2.379367285) (244, 2.622462192) (249, 2.619092885) (254, 2.631228959) (259, 2.776544601) (264, 2.79506289) (269, 2.780803463) (274, 2.948682958) (279, 3.091380529) (284, 3.242673435) (289, 3.326272413) (294, 3.427331551) (299, 3.243231825) (304, 3.491747066) (309, 3.792985006) (314, 3.809037269) (319, 3.807681672) (324, 3.791529428) (329, 3.702800021) (334, 3.867380047) (339, 3.66227304) (344, 3.967932055) (349, 4.047742247) (354, 4.175580924) (359, 4.149243785) (364, 4.066857203) (369, 4.236828666) (374, 4.291304352) (379, 4.487608756) (384, 4.804467428) (389, 5.043797328) (394, 5.262227729) (399, 5.642893598) (404, 5.65145953) (409, 6.018740658) (414, 6.135929588) (419, 6.550540235) (424, 6.7679592) (429, 6.637765334) (434, 6.82710062) (439, 6.973035556) (444, 6.992080279) (449, 6.858268223) (454, 7.149528018) (459, 7.232098787) (464, 7.152284012) (469, 7.189441424) (474, 7.311025211) (479, 7.474371372) (484, 7.433215638) (489, 7.48823332) (494, 7.384998351) (499, 7.387248613)
                };
            \addlegendentry{No Bonus Progress}

            \addplot[
                thick,
                orange!70!white
            ] coordinates {
                    (4, 0) (9, 0) (14, 0) (19, 0) (24, 0) (29, 0) (34, 0) (39, 0.045) (44, 0.128250001) (49, 0.169012501) (54, 0.178760625) (59, 0.151946531) (64, 0.158154553) (69, 0.143531198) (74, 0.236101518) (79, 0.316686291) (84, 0.444283346) (89, 0.610140843) (94, 0.658619716) (99, 0.627827759) (104, 0.698653609) (109, 0.673855568) (114, 0.722777233) (119, 0.960560648) (124, 0.966476551) (129, 0.876504858) (134, 0.795029129) (139, 0.831474759) (144, 0.856753545) (149, 0.878240513) (154, 0.876504436) (159, 0.875028771) (164, 0.826274456) (169, 0.932333288) (174, 1.042483293) (179, 0.937310797) (184, 0.918314178) (189, 0.980167051) (194, 1.026142035) (199, 0.997420731) (204, 1.062807621) (209, 1.123386497) (214, 1.118478521) (219, 1.108606743) (224, 1.167315731) (229, 1.185318371) (234, 1.347270613) (239, 1.657280022) (244, 1.883687819) (249, 2.144634648) (254, 1.872939451) (259, 2.177098532) (264, 2.200533752) (269, 2.350453689) (274, 2.527885636) (279, 2.571702791) (284, 2.765947372) (289, 2.851055165) (294, 2.888396889) (299, 3.129137357) (304, 3.309766753) (309, 3.388501741) (314, 3.480226479) (319, 3.591992501) (324, 3.553193125) (329, 3.522814156) (334, 3.744392283) (339, 3.789733441) (344, 3.851273423) (349, 3.883582409) (354, 3.955945048) (359, 3.826303291) (364, 4.072357795) (369, 4.111504126) (374, 4.124778707) (379, 4.246061501) (384, 4.539152376) (389, 4.448279519) (394, 4.879337592) (399, 4.992437153) (404, 4.868171579) (409, 5.336645842) (414, 5.436348965) (419, 5.485796621) (424, 5.538177128) (429, 5.767250559) (434, 6.031612975) (439, 6.476871029) (444, 6.729540376) (449, 6.800509319) (454, 7.000432821) (459, 7.410367896) (464, 7.718812711) (469, 8.060790804) (474, 8.051671683) (479, 8.343904929) (484, 7.970319189) (489, 8.543271211) (494, 8.883880529) (499, 9.084649449)
                };
            \addlegendentry{No Game Progress}

        \end{axis}
    \end{tikzpicture}
    \caption{Upper bonus achievement rate (EMA) by feature ablation}
    \label{fig:feature-ablation-bonus}
\end{figure}
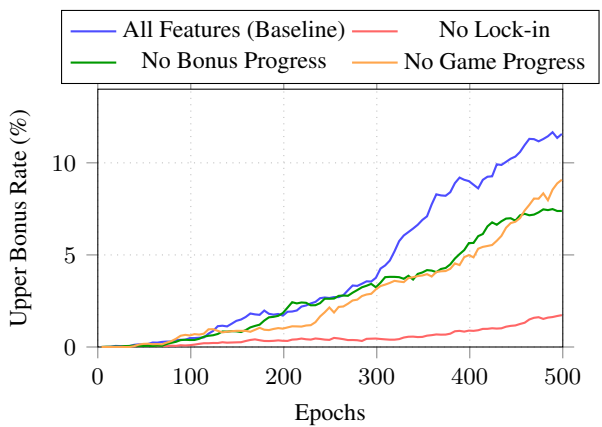

While a number of additional representational choices were explored, one of the most important is the state representation of the dice.

First, we ablated the basic representation (bin count vs one-hot) to understand their impact.
While the network can theoretically learn to reconstruct either representation from the other, in practice we found that using both improved reliability, as demonstrated in Figure~\ref{fig:dice-representation-bonus}.

For the full-game model, we added several additional features to the state representation: $\phi_{\mathrm{progress}}(t)$ and $\phi_{\mathrm{potential}}(\mathbf{d}, \mathbf{c})$
while reusing the same underlying neural network architecture as the single-turn model.
We intentionally omitted the $\phi_{\mathrm{potential}}(\mathbf{d}, \mathbf{c})$ feature in single turn, as we wanted to ensure the model was capable of learning category potentials.
To understand the importance of each of these features, they were ablated individually, with results shown in Figure~\ref{fig:feature-ablation-bonus}.

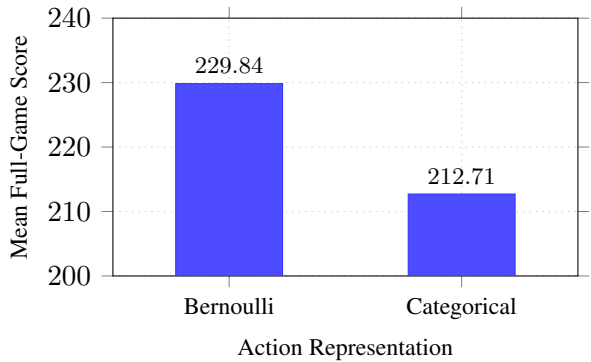
\begin{figure}[t]
    \centering
    \begin{tikzpicture}
        \begin{axis}[
                ybar,
                width=\columnwidth,
                height=5cm,
                xlabel={Action Representation},
                ylabel={Mean Full-Game Score},
                symbolic x coords={Bernoulli, Categorical},
                xtick=data,
                xticklabel style={font=\small},
                ylabel style={font=\small},
                xlabel style={font=\small},
                bar width=40pt,
                ymin=200, ymax=240,
                grid=both,
                grid style={dotted},
                tick align=outside,
                nodes near coords,
                nodes near coords style={font=\footnotesize, anchor=south},
                enlarge x limits=0.5,
            ]

            \addplot[
                fill=blue!70!white,
                draw=blue!80,
                error bars/.cd,
                y dir=both,
                y explicit
            ] coordinates {
                    (Bernoulli, 229.84)
                    (Categorical, 212.71)
                };

        \end{axis}
    \end{tikzpicture}
    \caption{Performance comparison: Bernoulli vs Categorical action representation}
    \label{fig:bernoulli-vs-categorical}
\end{figure}

\begin{figure}[b]
    \centering
    \begin{tikzpicture}
        \begin{axis}[
                width=\columnwidth,
                height=6cm,
                xlabel={Epochs},
                ylabel={Yahtzee Rate (\%)},
                xmin=0, xmax=500,
                ymin=0, ymax=50,
                grid=both,
                grid style={dotted},
                tick align=outside,
                tick label style={font=\footnotesize},
                label style={font=\footnotesize},
                legend style={at={(0.98,0.98)},anchor=north east,font=\footnotesize}
            ]

            \addplot[
                thick,
                blue!70!white
            ] coordinates {
                    (4, 0.2) (9, 0.1) (14, 0.1) (19, 0.1) (24, 0.2) (29, 0.2) (34, 0.2) (39, 0.2) (44, 0.1) (49, 1.6) (54, 1) (59, 1) (64, 1.4) (69, 1.7) (74, 1) (79, 2) (84, 1.8) (89, 2.5) (94, 1) (99, 1.9) (104, 1) (109, 1.7) (114, 1.3) (119, 1.5) (124, 1.3) (129, 2.2) (134, 3.4) (139, 2.3) (144, 2.1) (149, 2.6) (154, 2.6) (159, 3.2) (164, 4.2) (169, 3.9) (174, 2.9) (179, 3.4) (184, 3.4) (189, 3.1) (194, 3.6) (199, 4.9) (204, 3.8) (209, 4.7) (214, 4.9) (219, 5.1) (224, 3.7) (229, 5.6) (234, 5.1) (239, 4.1) (244, 6.2) (249, 6.8) (254, 5.9) (259, 6.8) (264, 7.6) (269, 14.3) (274, 18.5) (279, 20.4) (284, 22.8) (289, 24.4) (294, 23) (299, 23.6) (304, 23.2) (309, 23.8) (314, 24.1) (319, 22.8) (324, 25.3) (329, 25.5) (334, 26.7) (339, 25) (344, 23.9) (349, 27.4) (354, 29.9) (359, 29.5) (364, 28) (369, 28.4) (374, 27.3) (379, 28.6) (384, 29.2) (389, 29) (394, 32.7) (399, 32.6) (404, 32.9) (409, 30.4) (414, 32) (419, 31.7) (424, 35.1) (429, 33.6) (434, 33.2) (439, 31.4) (444, 32.2) (449, 30.2) (454, 33.7) (459, 34.5) (464, 31.5) (469, 34.3) (474, 31.9) (479, 31.8) (484, 32.6) (489, 31) (494, 37.4) (499, 32)
                };
            \addlegendentry{Categorical}

            \addplot[
                thick,
                red!60!white
            ] coordinates {
                    (4, 0) (9, 0) (14, 0.2) (19, 0) (24, 0.1) (29, 0) (34, 0.3) (39, 1.2) (44, 0) (49, 0.5) (54, 0.4) (59, 0.7) (64, 0.5) (69, 0.8) (74, 0.7) (79, 0.7) (84, 0.7) (89, 0.7) (94, 1.4) (99, 0.7) (104, 1.2) (109, 2.5) (114, 0.9) (119, 0.4) (124, 0.9) (129, 1.3) (134, 2) (139, 0.4) (144, 0.4) (149, 1.3) (154, 0.5) (159, 0.9) (164, 0.3) (169, 0.8) (174, 1.2) (179, 1.4) (184, 1.1) (189, 1.6) (194, 1.6) (199, 1.2) (204, 1) (209, 1.1) (214, 1.4) (219, 2.1) (224, 1.5) (229, 2.9) (234, 2.1) (239, 2.8) (244, 2.9) (249, 3.2) (254, 4.4) (259, 2.2) (264, 1.5) (269, 3.2) (274, 2.4) (279, 1.7) (284, 2.3) (289, 2.3) (294, 2.8) (299, 1.7) (304, 2.9) (309, 2.3) (314, 3.5) (319, 4) (324, 4.1) (329, 4.2) (334, 5) (339, 6.1) (344, 3.4) (349, 3.1) (354, 3.9) (359, 4.2) (364, 3.5) (369, 5.7) (374, 3.8) (379, 4.1) (384, 3.4) (389, 4.3) (394, 7) (399, 4.6) (404, 5) (409, 5.3) (414, 5.1) (419, 5.1) (424, 4.9) (429, 7.4) (434, 6) (439, 5.4) (444, 6.5) (449, 5.3) (454, 5.4) (459, 5.5) (464, 5.6) (469, 7.4) (474, 7) (479, 5.4) (484, 4.7) (489, 5.1) (494, 5.2) (499, 5.6)
                };
            \addlegendentry{Bernoulli}

        \end{axis}
    \end{tikzpicture}
    \caption{Learning Yahtzee with different action representations}
    \label{fig:action-yahtzee-learning}
\end{figure}
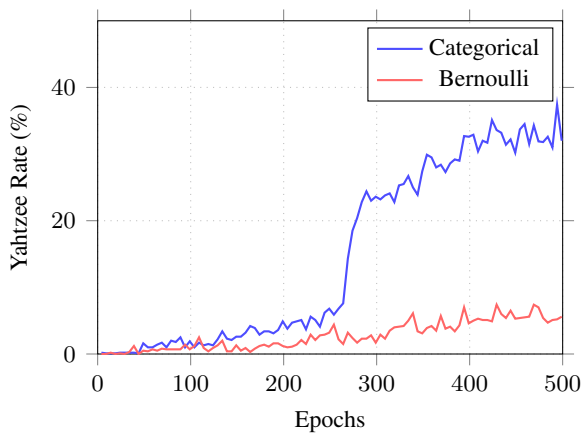

Lastly, we tested our hypothesis that a 32-way categorical would prove beneficial to complex actions that required specific combinations of dice to be held (see Section \ref{sec:rolling-action}).
Figure~\ref{fig:bernoulli-vs-categorical} shows the performance comparison, while Figure~\ref{fig:action-yahtzee-learning} illustrates how each representation learns to achieve Yahtzee during training.

\subsubsection{Architectural Ablations}
\label{sec:arch-ablations}

\begin{figure}[t]
    \centering
    \begin{tikzpicture}
        \begin{axis}[
                width=\columnwidth,
                height=0.5\textwidth,
                xlabel={Network Width (Hidden Units)},
                ylabel={Network Depth (Hidden Layers)},
                colormap/viridis,
                xlabel style={font=\small},
                ylabel style={font=\small},
                title style={font=\small},
                xtick={1,2,3},
                xticklabels={300, 600, 900},
                ytick={1,2,3},
                yticklabels={2, 3, 4},
                xticklabel style={font=\small},
                yticklabel style={font=\small},
                enlargelimits=0.3,
                point meta min=185,
                point meta max=235,
                nodes near coords,
                nodes near coords style={font=\small, color=black},
                every node near coord/.append style={xshift=0pt, yshift=0pt},
            ]

            \addplot[
                matrix plot,
                mesh/cols=3,
                point meta=explicit
            ] table[meta=score] {
                    x y score
                    1 1 203.959
                    2 1 227.618
                    3 1 230.508
                    1 2 194.403
                    2 2 205.788
                    3 2 212.495
                    1 3 192.501
                    2 3 190.864
                    3 3 187.407
                };

        \end{axis}
    \end{tikzpicture}
    \caption{Network size ablation}
    \label{fig:architecture-heatmap}
\end{figure}
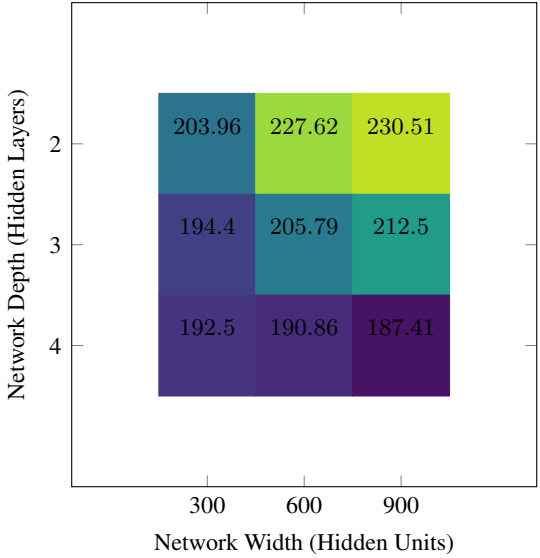

We performed a simple grid search ablation to understand if our chosen architecture of 3 hidden layers of 600 units each was optimal.
Yahtzee is a fairly complex game, so we expected shorter, but wider networks to perform best. Note that each of these has a different number of total parameters,
so this is not a pure ablation of depth vs. width.
Results are shown in Figure~\ref{fig:architecture-heatmap}.

Based on \cite{bjorck-2022-high-variance}, we hypothesized that layer normalization \cite{ba-2016-layernorm} would improve training stability and performance
and used it in all of our main experiments. This was ablated to understand its true impact, with learning curves compared in Figure~\ref{fig:layernorm-learning}.

\input{figures/layernorm}



\subsubsection{Credit Assignment: TD(0) vs GAE}

Later in this research, we noticed the main issue with our network was that it was struggling to earn the bonus, learning it very slowly.
We first hypothesized that this was due to high variance in REINFORCE, so we switched to A2C with TD(0) targets. However, the issue
persisted. We then hypothesized that the TD(0) targets were not providing sufficient credit assignment for the long-term bonus reward,
so we switched to GAE with various $\lambda$ values to understand if this would help.

Unfortunately, we found that GAE did not improve performance over TD(0), and values that were too high ($\lambda \geq 0.8$)
significantly degraded performance, as shown in Figures~\ref{fig:gae-lambda-performance} and~\ref{fig:gae-lambda-bonus}.

\begin{figure}[t]
    \centering
    \begin{tikzpicture}
        \begin{axis}[
                ybar,
                width=\columnwidth,
                height=5cm,
                xlabel={$\lambda$ (GAE Parameter)},
                ylabel={Mean Full-Game Score},
                symbolic x coords={0.0, 0.2, 0.4, 0.6, 0.8, 0.9, 0.95},
                xtick=data,
                xticklabel style={font=\small},
                ylabel style={font=\small},
                xlabel style={font=\small},
                bar width=15pt,
                ymin=190, ymax=235,
                grid=both,
                grid style={dotted},
                tick align=outside,
                nodes near coords,
                nodes near coords style={font=\footnotesize, anchor=south},
            ]

            \addplot[
                fill=blue!70!white,
                draw=blue!80!white,
                error bars/.cd,
                y dir=both,
                y explicit
            ] coordinates {
                    (0.0, 227.0) +- (0, 0)
                    (0.2, 229.1) +- (0, 0)
                    (0.4, 228.5) +- (0, 0)
                    (0.6, 224.1) +- (0, 0)
                    (0.8, 213.8) +- (0, 0)
                    (0.9, 202.1) +- (0, 0)
                    (0.95, 196.7) +- (0, 0)
                };

        \end{axis}
    \end{tikzpicture}
    \caption{Final performance by GAE lambda}
    \label{fig:gae-lambda-performance}
\end{figure}
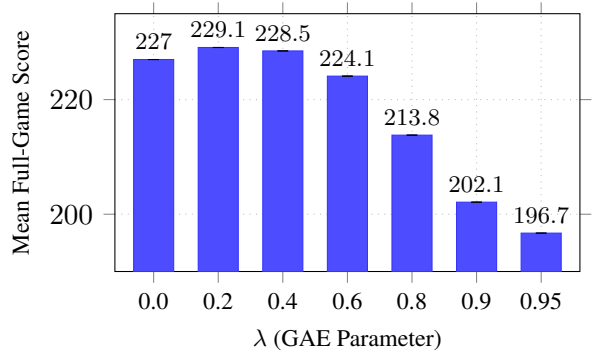
\begin{figure}[H]
    \centering
    \begin{tikzpicture}
        \begin{axis}[
                ybar,
                width=\columnwidth,
                height=5cm,
                xlabel={$\lambda$ (GAE Parameter)},
                ylabel={Upper Bonus Rate (\%)},
                symbolic x coords={0.0, 0.2, 0.4, 0.6, 0.8, 0.9, 0.95},
                xtick=data,
                xticklabel style={font=\small},
                ylabel style={font=\small},
                xlabel style={font=\small},
                bar width=15pt,
                ymin=0, ymax=13,
                grid=both,
                grid style={dotted},
                tick align=outside,
                nodes near coords,
                nodes near coords style={font=\footnotesize, anchor=south},
            ]

            \addplot[
                fill=blue!70!white,
                draw=blue!80!white,
                error bars/.cd,
                y dir=both,
                y explicit
            ] coordinates {
                    (0.0, 11.3)
                    (0.2, 10.4)
                    (0.4, 10.6)
                    (0.6, 10.3)
                    (0.8, 8.7)
                    (0.9, 5.3)
                    (0.95, 4.8)
                };

        \end{axis}
    \end{tikzpicture}
    \caption{Upper bonus achievement by GAE lambda}
    \label{fig:gae-lambda-bonus}
\end{figure}
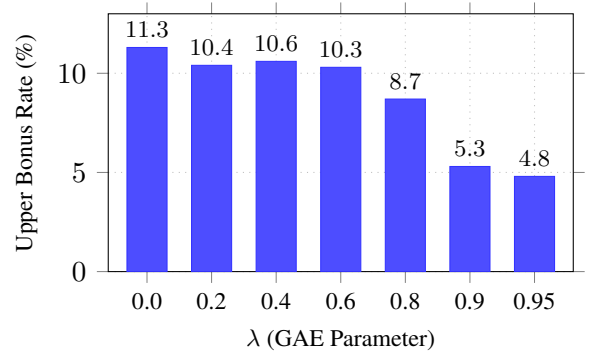

\subsubsection{Entropy Sensitivity}
\begin{table}[H]
    \centering
    \small
    \caption{Entropy regime definitions}
    \label{tab:entropy-regime-definitions}
    \begin{tabular}{lcccc}
        \hline
        \textbf{Regime} & $\beta_{\mathrm{roll}}$ & $\beta_{\mathrm{score}}$ & Hold / Anneal \\
        \hline
        None            & 0.0                     & 0.0                      & 0 / 0         \\
        Low             & 0.05 $\rightarrow$ 0.01 & 0.01 $\rightarrow$ 0.005 & 0.2 / 0.4     \\
        Baseline        & 0.1 $\rightarrow$ 0.02  & 0.03 $\rightarrow$ 0.01  & 0.3 / 0.6     \\
        High            & 0.2 $\rightarrow$ 0.04  & 0.06 $\rightarrow$ 0.02  & 0.35 / 0.65   \\
        \hline
    \end{tabular}
\end{table}

During experimentation, we noticed that the entropy regularization coefficients had a significant impact on training stability and final performance,
as described in Section~\ref{sec:entropy}.
To better understand this sensitivity, we trained models under three different entropy regimes: Low Entropy, Baseline, and High Entropy, as defined in Table~\ref{tab:entropy-regime-definitions}.
The learning curves and entropy values are shown in Figure~\ref{fig:entropy-analysis}.

\input{figures/entropy_mean_score}

\subsubsection{Reward Shaping}
\begin{figure}[b!]
    \centering
    \begin{tikzpicture}
        \begin{axis}[
                width=\columnwidth,
                height=\columnwidth,      
                axis equal image,
                xmin=0.5, xmax=4.5,       
                ymin=0.5, ymax=4.5,
                enlargelimits=false,
                xlabel={Shaping loss weight},
                ylabel={Shaping reward weight},
                colormap/viridis,
                xlabel style={font=\small},
                ylabel style={font=\small},
                title style={font=\small},
                xtick={1,2,3,4},
                xticklabels={0.25, 0.5, 1.0, 2.0},
                ytick={1,2,3,4},
                yticklabels={0.25, 0.5, 1.0, 2.0},
                xticklabel style={font=\small},
                yticklabel style={font=\small},
                point meta min=224,
                point meta max=233,
                nodes near coords,
                nodes near coords style={font=\small, color=black},
                every node near coord/.append style={xshift=0pt, yshift=0pt},
            ]

            \addplot[
                matrix plot,
                mesh/cols=4,
                point meta=explicit
            ] table[row sep=\\,meta=score] {
                    x  y  score \\
                    1  1  225.4 \\
                    1  2  230.7 \\
                    1  3  227.4 \\
                    1  4  224.6 \\
                    2  1  228.1 \\
                    2  2  228.3 \\
                    2  3  227.3 \\
                    2  4  229.7 \\
                    3  1  227.8 \\
                    3  2  229.2 \\
                    3  3  232.7 \\
                    3  4  231.5 \\
                    4  1  229.2 \\
                    4  2  228.5 \\
                    4  3  232.4 \\
                    4  4  232.2 \\
                };

        \end{axis}
    \end{tikzpicture}
    \caption{Shaping parameters ablation}
    \label{fig:shaping-heatmap}
\end{figure}
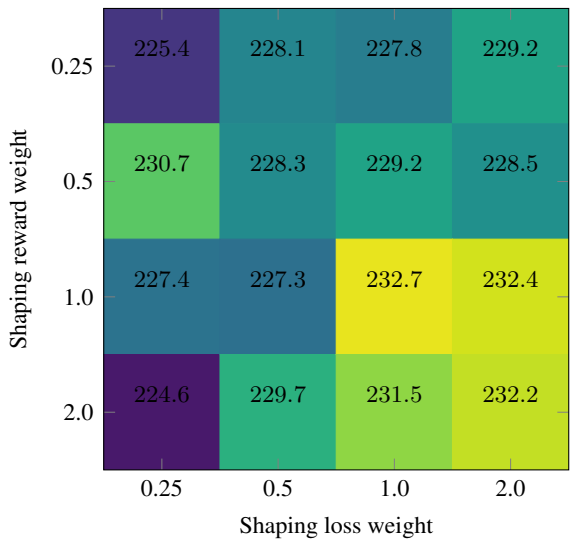
We co-varied the shaping loss weight and the strength of the shaping reward
to understand their impact on final performance. Figure~\ref{fig:shaping-heatmap} shows the results of this ablation study.

\subsubsection{Summary}
In summary, we found that A2C with TD(0) targets, a combined dice representation, Layer Normalization,
and carefully tuned entropy regularization produced the best results.
Table~\ref{tab:full-game-summary} presents a comprehensive comparison of all algorithms tested.

\begin{table*}[htb!]
    \centering
    \small
    \caption{Full-game performance summary}
    \label{tab:full-game-summary}
    \begin{tabular}{lcccccc}
        \hline
        \textbf{Algorithm}         & \textbf{Training Budget} & \textbf{Mean Score} & \textbf{Std Dev} & \textbf{Bonus Rate (\%)} & \textbf{Yahtzee Rate (\%)} & \textbf{$\geq$250 (\%)} \\
        \hline
        DP Optimal                 & --                       & 254.59              & --               & 68.12\%                  & 33.74\%                    & 48.37\%                 \\
        \hline
        A2C                        & 250K games               & 230.38              & 2503.83          & 11.37\%                  & 31.08\%                    & 27.82\%                 \\
        A2C                        & 1M games                 & 241.78              & 3230.86          & 24.93\%                  & 34.05\%                    & 36.87\%                 \\
        PPO ($\lambda=0.3$, $k=5$) & 50k games                & 204.54              & 860.02           & 2.49\%                   & 6.54\%                     & 6.08\%                  \\
        PPO ($\lambda=0.3$, $k=4$) & 250K games               & 230.20              & 2345.5           & 19.71\%                  & 24.87\%                    & 28.38\%                 \\
        REINFORCE (full-game)      & 250K games               & 189.84              & 812.24           & 2.83\%                   & 1.70\%                     & 2.06\%                  \\
        REINFORCE (full-game)      & 1M games                 & 203.73              & 1265.83          & 6.44\%                   & 10.61\%                    & 9.22\%                  \\
        REINFORCE (single-turn)    & 500K games               & 201.48              & 1366.00          & 0.89\%                   & 19.63\%                    & 9.34\%                  \\
        \hline
    \end{tabular}
\end{table*}

For our best configuration, A2C trained over 1 million games, the final score distribution is compared to DP-optimal in Table \ref{tab:dp-score-distribution}.
\begin{table}[htb!]
    \centering
    \small
    \caption{$P(\text{score} \geq n)$, 100,000 games}
    \label{tab:dp-score-distribution}
    \begin{tabular}{rcc}
        \hline
        \textbf{$n$} & A2C      & DP                        \\
        \hline
        50           & 1.000000 & 1.000000                  \\
        100          & 0.999980 & 0.999998                  \\
        150          & 0.989730 & 0.991230                  \\
        200          & 0.820980 & 0.863584                  \\
        250          & 0.368730 & 0.483683                  \\
        300          & 0.109080 & 0.143265                  \\
        400          & 0.025960 & 0.038351                  \\
        500          & 0.004870 & 0.007192                  \\
        750          & 0        & $5.11603 \cdot 10^{-6}$   \\
        1000         & 0        & $5.57508 \cdot 10^{-9}$   \\
        1250         & 0        & $ 6.49213 \cdot 10^{-13}$ \\
        1500         & 0        & $ 3.93308 \cdot 10^{-19}$ \\
        \hline
    \end{tabular}
\end{table}

\subsection{Policy Analysis}
\subsubsection{Category Usage}
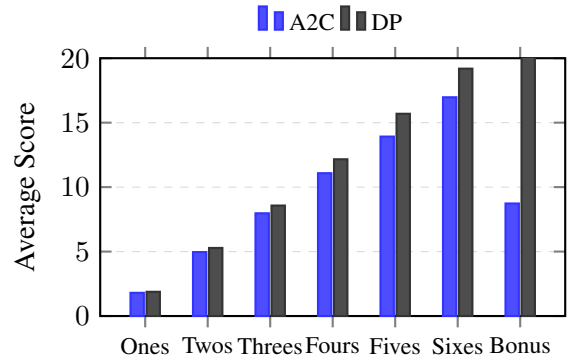
\begin{figure}[htb!]
    \centering
    \begin{tikzpicture}
        \begin{axis}[
                ybar=1pt,
                width=\columnwidth,
                height=5cm,
                bar width=5pt,
                symbolic x coords={Ones,Twos,Threes,Fours,Fives,Sixes,Bonus},
                xtick=data,
                xticklabel style={font=\small},
                ylabel={Average Score},
                ymin=0,
                ymax=20,
                ymajorgrids=true,
                grid style={dashed,gray!30},
                legend style={
                        at={(0.5,1.05)},
                        anchor=south,
                        legend columns=4,
                        font=\footnotesize,
                        draw=none,
                        fill=none
                    },
                enlarge x limits=0.12,
                axis line style={thick},
                tick style={thick},
            ]

            \addplot[fill=blue!70!white, draw=blue!80!white, thick] coordinates {
                    (Ones,   1.80) (Twos,   4.96) (Threes, 7.97) (Fours,  11.08)
                    (Fives,  13.92) (Sixes,  16.97) (Bonus,  8.73)
                };



            \addplot[fill=black!70, draw=black!80, thick] coordinates {
                    (Ones,   1.88) (Twos,   5.28) (Threes, 8.57) (Fours,  12.16)
                    (Fives,  15.69) (Sixes,  19.19) (Bonus,  23.84)
                };

            \legend{A2C,DP}

        \end{axis}
    \end{tikzpicture}
    \caption{Upper section and bonus scores}
    \label{fig:category-upper}
\end{figure}

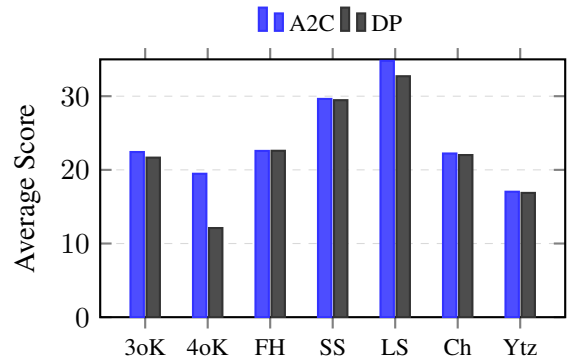
\begin{figure}[htb!]
    \centering
    \begin{tikzpicture}
        \begin{axis}[
                ybar=1pt,
                width=\columnwidth,
                height=5cm,
                bar width=5pt,
                symbolic x coords={3oK,4oK,FH,SS,LS,Ch,Ytz},
                xtick=data,
                xticklabel style={font=\small},
                ylabel={Average Score},
                ymin=0,
                ymax=35,
                ymajorgrids=true,
                grid style={dashed,gray!30},
                legend style={
                        at={(0.5,1.05)},
                        anchor=south,
                        legend columns=4,
                        font=\footnotesize,
                        draw=none,
                        fill=none
                    },
                enlarge x limits=0.12,
                axis line style={thick},
                tick style={thick},
            ]

            \addplot[fill=blue!70!white, draw=blue!80!white, thick] coordinates {
                    (3oK,  22.42) (4oK,  19.47) (FH,   22.57) (SS,   29.63)
                    (LS,   34.80) (Ch,   22.22) (Ytz,  17.03)
                };



            \addplot[fill=black!70, draw=black!80, thick] coordinates {
                    (3oK,  21.66) (4oK,  12.10) (FH,   22.59) (SS,   29.46)
                    (LS,   32.71) (Ch,   22.01) (Ytz,  16.87)
                };

            \legend{A2C,DP}

        \end{axis}
    \end{tikzpicture}
    \caption{Lower section scores}
    \label{fig:category-lower}
\end{figure}

To understand the overall performance of the A2C agent, we compare its average scores in each category against the relevant DP-optimal score
in figures \ref{fig:category-upper} and \ref{fig:category-lower}.

\subsubsection{Strategy Comparison Across Agents}
We also compared some high-level strategy metrics across our different agents to understand how their learned policies differed.
Table~\ref{tab:top-categories-by-turn} shows the top three most frequently used categories at each turn of the game, providing insights into the agent's strategy throughout a game.

\input{figures/strategy_fill_turn}

\begin{table*}[tb]
    \centering
    \small
    \caption{Top 3 most frequently used categories by turn (A2C, 1M games, 100K evaluation games)}
    \label{tab:top-categories-by-turn}
    \begin{tabular}{clcr}
        \hline
        \textbf{Turn} & \textbf{Category} & \textbf{Usage \%} & \textbf{Median Score} \\
        \hline
        1             & Small Straight    & 17.0\%            & 30.0                  \\
                      & Full House        & 12.3\%            & 25.0                  \\
                      & Large Straight    & 11.8\%            & 40.0                  \\
        \hline
        2             & Small Straight    & 15.1\%            & 30.0                  \\
                      & Full House        & 11.2\%            & 25.0                  \\
                      & Large Straight    & 10.5\%            & 40.0                  \\
        \hline
        3             & Small Straight    & 12.8\%            & 30.0                  \\
                      & Full House        & 10.4\%            & 25.0                  \\
                      & Large Straight    & 9.3\%             & 40.0                  \\
        \hline
        4             & Small Straight    & 10.8\%            & 30.0                  \\
                      & Full House        & 9.9\%             & 25.0                  \\
                      & Fours             & 8.9\%             & 12.0                  \\
        \hline
        5             & Small Straight    & 9.1\%             & 30.0                  \\
                      & Full House        & 9.0\%             & 25.0                  \\
                      & Fours             & 8.8\%             & 12.0                  \\
        \hline
        6             & Twos              & 9.3\%             & 5.0                   \\
                      & Fours             & 9.1\%             & 12.0                  \\
                      & Threes            & 9.0\%             & 9.0                   \\
        \hline
        7             & Twos              & 10.3\%            & 4.0                   \\
                      & Ones              & 9.3\%             & 1.0                   \\
                      & Threes            & 9.2\%             & 9.0                   \\
        \hline
        8             & Ones              & 11.2\%            & 1.0                   \\
                      & Twos              & 11.0\%            & 4.0                   \\
                      & Three of a Kind   & 9.7\%             & 23.0                  \\
        \hline
        9             & Ones              & 12.8\%            & 1.0                   \\
                      & Twos              & 11.0\%            & 4.0                   \\
                      & Three of a Kind   & 9.4\%             & 23.0                  \\
        \hline
        10            & Ones              & 13.9\%            & 1.0                   \\
                      & Twos              & 10.2\%            & 4.0                   \\
                      & Chance            & 9.4\%             & 22.0                  \\
        \hline
        11            & Ones              & 13.8\%            & 1.0                   \\
                      & Yahtzee           & 13.0\%            & 0.0                   \\
                      & Twos              & 9.3\%             & 4.0                   \\
        \hline
        12            & Yahtzee           & 20.5\%            & 0.0                   \\
                      & Ones              & 10.9\%            & 2.0                   \\
                      & Four of a Kind    & 8.0\%             & 6.0                   \\
        \hline
        13            & Yahtzee           & 26.6\%            & 0.0                   \\
                      & Large Straight    & 13.4\%            & 0.0                   \\
                      & Four of a Kind    & 11.7\%            & 0.0                   \\
        \hline
    \end{tabular}
\end{table*}

\section{Discussion}

\subsection{Summary}
In this work, we attempted to use policy-gradient reinforcement learning methods to teach agents to play \textit{Yahtzee} through self-play.
We found that with appropriate algorithmic and architectural choices, it is possible to approach near-optimal performance.
Advantage Actor-Critic (A2C) with TD(0) was consistently stable and efficient; REINFORCE and PPO were more fragile and underperformed at equal training budgets.
Our best A2C agent, trained over 1 million games, achieved a median score of 241.78 points over 100,000 evaluation games, which is within 5.0\% of the DP-optimal score of 254.59.
These results are best reported in Table~\ref{tab:full-game-summary}.

We found that single-turn REINFORCE gets surprisingly high scores,
often outperforming full-game REINFORCE agents, but fails to learn a coherent bonus strategy.
We observed a tradeoff between single-turn and full-game performance, especially at higher performance levels. as seen in Figure~\ref{fig:pareto-frontier},
and found a Pareto frontier between the two objectives.

Our ablation studies highlighted several design choices that significantly impacted final performance.
The choice of RL algorithm and credit assignment was crucial; A2C with TD(0) outperformed both PPO and REINFORCE in terms of stability and final score,
as shown in Table~\ref{tab:full-game-summary}.
As seen in Figures~\ref{fig:dice-representation-bonus} and~\ref{fig:feature-ablation-bonus}, the state and action encodings played a significant role; specifically providing (rather than forcing the network to learn) some easily calculable features improved learning.
Categorical action distributions outperformed Bernoulli ones, as seen in Figure~\ref{fig:bernoulli-vs-categorical}, allowing the network to better model compound actions.
LayerNorm improved stability and final performance, as shown in Figure~\ref{fig:layernorm-learning}.
Lastly, we found diminishing returns for model size; larger models improved performance up to a point, but after a certain size, gains were minimal, as shown in Figure~\ref{fig:architecture-heatmap}.

Moving from single-turn to full-game play, REINFORCE struggled with the
variance and credit assignment challenges. Performance immediately improved when we switched to TD(0) returns and A2C,
after adjusting certain hyperparameters (notably the critic coefficient).
When A2C still struggled with the upper bonus strategy, we implemented GAE returns,
However, they did not lead to significant improvement, as shown in Figures~\ref{fig:gae-lambda-performance} and~\ref{fig:gae-lambda-bonus};
moderate values of GAE were slightly helpful, but high values led back to instability.

Entropy regularization played a huge role in stabilizing training and encouraging exploration, best seen in Figure~\ref{fig:entropy-analysis};
striking the right balance of explore vs exploit was critical. There's a narrow sweet spot:
too little entropy and the policy collapses, does not explore and we get stuck in local minima;
too much entropy and the policy becomes too random to learn important strategies.

There were some limitations to our approach. First, we fixed the training budget to 1 million games for all agents,
but could only explore the hyperparameter space using 250,000 games per run.
Second, we typically ran only a single seed per configuration due to compute constraints, so it's possible that some results were affected by random chance.
Third, we only explored a limited set of architectures and hyperparameters; more extensive sweeps, especially around PPO,
could yield better performance.
We also did not explore transfer learning from single-turn to full-game agents, which could improve sample efficiency.

\subsection{Strategy \& Failure Modes}
The primary differentiators between our RL agents and the DP optimal solution are in the upper section (and bonus), four-of-a-kind and Yahtzee.
The high performance for many agents on Yahtzee category was interesting, as it requires agents to be performing at a competent level across multiple turns.
The interesting takeaway here is that agents appear to
exhibit a "mode shift" in their strategy once they figure out Yahtzee (as shown in Figure~\ref{fig:action-yahtzee-learning}),
whereas the bonus is learned more gradually over time (as shown in Figure~\ref{fig:dice-representation-bonus}).

From the per-turn statistics in Table~\ref{tab:top-categories-by-turn}, we can infer a typical game,
which begins with the agent locking in straights or full houses.
These are fixed, high-value categories that provide a strong foundation. The agent also prioritizes the 4-of-a-kind category early on,
typically on the 6th turn. The agent then turns its attention to the upper section, often scoring in the
4's, 5's, 6's, sometimes taking 3-of-a-kind instead.
However, it seems to prefer taking a lower-section score for 5's and 6's rather than using them to build towards the bonus.
The agent seems to understand the importance of reaching the 63-point threshold for the bonus,
but perhaps does not prioritize it early enough in the game.
We also see Ones, Twos, and Chance being used as "dump" categories later in the game when no better options are available.
The agent does follow common wisdom, avoiding zeroing out high-value categories until forced to late in the game.

It could be that the agent is risk-averse, preferring the immediate points from lower-section categories, or it simply doesn't properly
realize it needs to take at least one above-average score in each of the 4's, 5's, and 6's to offset lower scores in the 1's, 2's, and 3's later in the game.
This is the core issue for the agent; it has difficulty learning when to pivot to bonus-seeking behavior later in training.
We see evidence of this when comparing the category mean scores against DP in Figures~\ref{fig:category-upper} and~\ref{fig:category-lower}.
Fundamentally, the agent needs to learn to sacrifice the marginal points from the 5th dice in 4-of-a-kind, in order to put itself in a better position to earn the bonus later.

The agent does surprisingly well in achieving Yahtzee, indicating it has figured out how to lock in on selecting pairs, triples, and quads when the opportunity arises.
It would be worth investigating further if the agent is explicitly targeting Yahtzee or if it's a byproduct of its general strategy,
and if the agent recognizes that even 1's and 2's can be worth a lot, if they score a Yahtzee.

Reward shaping definitely helped the agent learn the bonus more quickly, but had its limits.
At very high levels, the agent completely derailed and failed to prioritize the lower section, most notably ignoring Yahtzees.
Since we are using a learned potential function, we do break the theoretical guarantees of potential-based reward shaping \cite{ng-1999-reward-shaping},
this could explained the mixed results.

In general our results backed up many known challenges in reinforcement learning:
deep RL methods struggle with long-horizons, credit assignment, variance,
and balanced exploration.
Models are typically very quick to learn local strategies that yield immediate rewards;
for example, our agents quickly snapped to the DP-optimal scores for full-house, straights, and even Yahtzee.
However, they systematically under optimized the only part of Yahtzee that require planning over the entire game: the upper section bonus.
Yahtzee exposes this dynamic in a compact setting, showing it's value as a benchmark for RL research.

\section{Conclusion and Future Work}

Learning a robust policy for \textit{Yahtzee} using reinforcement learning presents several interesting challenges and insights.
We showed that with appropriate algorithmic choices, it is possible to approach near-optimal performance using self-play alone.
Our results back up theoretical results in the literature regarding training stability and sample efficiency of common RL algorithms.
Our analysis of learned policies showed that these algorithms often struggle to learn rare, yet high-reward strategies, especially if they require strong coherence over longer time horizons.

Future research could be done to find architectures, samples, and learning methods that allow the model to better approximate optimal play, more efficiently.
Transfer learning could be explored further to see if knowledge from single-turn optimization could be effectively transferred to full-game, multiplayer Yahtzee, or other variants of the game.
For example, curriculum learning approaches, where the agent is gradually exposed to more complex scenarios over time, could be used to help the model overcome some challenges outlined in this paper.
For the multiplayer setting, future work could explore permutation-invariant architectures such as Deep Sets \cite{zaheer2018deepsets} or embeddings with self-attention to handle unsorted dice or opponent states\cite{vaswani-2017-attention}.
Additionally, \textrm{Yahtzee} could also be considered as a candidate environment for research into hierarchical reinforcement learning methods \cite{Barto2003}.

\bibliographystyle{acl_natbib}
\bibliography{csml_bibliography}

\clearpage
\appendix

\section{Reproducibility}
The custom gym environment, training code, models, and final evaluation statistics for this project are \href{https://github.com/papetronics/case-studies-final-project}{available on GitHub}.

Likewise, data for all experiments used in this graph is available in \href{https://api.wandb.ai/links/papetronics/qfovf68e}{this Weights \& Biases report}.

\section{Hyperparameters}
The following hyperparameters were used for the baseline models:

\begin{table}[H]
    \centering
    \small
    \caption{Shared hyperparameters across all algorithms}
    \label{tab:shared-hyperparameters}
    \begin{tabular}{lc}
        \hline
        \textbf{Hyperparameter}       & \textbf{Value} \\
        \hline
        $d_h$ (Hidden Size)           & 600            \\
        $L$ (Hidden Layers)           & 3              \\
        $p_d$ (Dropout Rate)          & 0.1            \\
        $B$ (Games per Batch)         & 20             \\
        Activation Function           & Swish          \\
        Rolling Action Representation & Categorical    \\
        \hline
    \end{tabular}
\end{table}

\begin{table}[H]
    \centering
    \small
    \setlength{\tabcolsep}{4pt}
    \caption{Algorithm-specific hyperparameters}
    \label{tab:algorithm-hyperparameters}
    \begin{tabular}{lccc}
        \hline
        \textbf{Hyperparameter}        & \textbf{REINFORCE} & \textbf{A2C} & \textbf{PPO} \\
        \hline
        $\alpha$                       & 0.001              & 0.0001       & 0.001        \\
        $r_{\alpha}$ (Min LR Ratio)    & 0.01               & 0.05         & 0.05         \\
        $\gamma$ (min)                 & 0.95               & 0.99         & 0.99         \\
        $\gamma$ (max)                 & 1.0                & 0.99         & 0.99         \\
        $\tau_{\mathrm{clip}}$         & 0.0                & 1.0          & 1.0          \\
        $\lambda_V$                    & 0.025              & 0.005        & 0.02         \\
        $\beta_{\mathrm{roll}}$ (max)  & 0.1                & 0.1          & 0.005        \\
        $\beta_{\mathrm{roll}}$ (min)  & 0.01               & 0.02         & 0.005        \\
        $\beta_{\mathrm{score}}$ (max) & 0.02               & 0.03         & 0.05         \\
        $\beta_{\mathrm{score}}$ (min) & 0.003              & 0.01         & 0.01         \\
        Entropy Hold Period            & 0.25               & 0.075        & 0.05         \\
        Entropy Anneal Period          & 0.91               & 0.9          & 0.9          \\
        \hline
    \end{tabular}
\end{table}

\begin{table}[H]
    \centering
    \small
    \caption{PPO-specific hyperparameters}
    \label{tab:ppo-hyperparameters}
    \begin{tabular}{lc}
        \hline
        \textbf{Hyperparameter} & \textbf{Value} \\
        \hline
        PPO Clip $\epsilon$     & 0.2            \\
        PPO Games per Minibatch & 4              \\
        PPO Epochs              & 4              \\
        \hline
    \end{tabular}
\end{table}

\section{Compute Costs}
Experiments were collected using a mix of a local RTX 3090 and AWS-hosted Tesla T4 GPUs.
The total cost of cloud compute was approximately \textbf{\$223}.
Over \textbf{481} training runs were logged in Weights \& Biases, totaling approximately \textbf{1,310.73 GPU hours}.

\section{AI Usage}

This paper utilized artificial intelligence tools in the following ways:
\begin{itemize}
    \item \textbf{GitHub Copilot (Claude Sonnet 4.5)} was used for typesetting assistance with LaTeX/KaTeX, IDE autocomplete suggestions during coding, and to occassionally perform straightforward refactorings, CUDA performance optimizations, and debugging.
    \item \textbf{ChatGPT (GPT-5.1)} was used for brainstorming ideas for reinforcement learning applications in games, guidance in hyperparameter tuning, helping to outline the structure of this paper, assistance in discovering relevant research and citations, and for writing tone and quality feedback.
\end{itemize}
All other content, including research methodology, analysis, results interpretation, and conclusions, represents original work by the author. The AI tools were not used to generate substantive content or analysis in this document.

\section{Yahtzee Scoring Rules}
\label{app:scoring}
Next we define the indicator functions for each of the scoring categories:

\begin{align*}
    \mathbb{I}_{3\mathrm{k}}(\mathbf{d})
     & = \mathbb{I}\bigl\{ \max_{v} n_v(\mathbf{d}) \ge 3 \bigr\} \\
    \mathbb{I}_{4\mathrm{k}}(\mathbf{d})
     & = \mathbb{I}\bigl\{ \max_{v} n_v(\mathbf{d}) \ge 4 \bigr\} \\
    \mathbb{I}_{\mathrm{full}}(\mathbf{d})
     & = \mathbb{I}\Bigl\{
    \exists i, j \in \{1, \mathellipsis, 6 \} \ \text{with} \ n_i(\mathbf{d}) = 3 \land n_j(\mathbf{d}) = 2
    \Bigr\}                                                       \\
    \mathbb{I}_{\mathrm{ss}}(\mathbf{d})
     & = \mathbb{I}\Bigl\{
    \exists k \in \{1,2,3\} \ \text{with}\
    \sum_{v=k}^{k+3} \mathbb{I}\{n_v(\mathbf{d}) > 0\} = 4
    \Bigr\}                                                       \\
    \mathbb{I}_{\mathrm{ls}}(\mathbf{d})
     & = \mathbb{I}\Bigl\{
    \exists k \in \{1,2\} \ \text{with}\
    \sum_{v=k}^{k+4} \mathbb{I}\{n_v(\mathbf{d}) > 0\} = 5
    \Bigr\}                                                       \\
    \mathbb{I}_{\mathrm{yahtzee}}(\mathbf{d})
     & = \mathbb{I}\bigl\{\max_v n_v(\mathbf{d}) = 5\bigr\}
\end{align*}

The potential score for each category can then be defined as:
\begin{align*}
    f_j(\mathbf{d})        & = j \cdot n_j(\mathbf{d}), \qquad j \in \{1,\dots,6\}                   \\
    f_7(\mathbf{d})        & = \mathbf{1}^\top \mathbf{d} \cdot \mathbb{I}_{3\mathrm{k}}(\mathbf{d}) \\
    f_8(\mathbf{d})        & = \mathbf{1}^\top \mathbf{d} \cdot \mathbb{I}_{4\mathrm{k}}(\mathbf{d}) \\
    f_9(\mathbf{d})        & = 25 \cdot \mathbb{I}_{\mathrm{full}}(\mathbf{d})                       \\
    f_{10}(\mathbf{d})     & = 30 \cdot \mathbb{I}_{\mathrm{ss}}(\mathbf{d})                         \\
    f_{11}(\mathbf{d})     & = 40 \cdot \mathbb{I}_{\mathrm{ls}}(\mathbf{d})                         \\
    f_{12}(\mathbf{d})     & = 50 \cdot \mathbb{I}_{\mathrm{yahtzee}}(\mathbf{d})                    \\
    f_{13}(\mathbf{d})     & = \mathbf{1}^\top \cdot \mathbf{d}                                      \\
    \mathbf{f}(\mathbf{d}) & =
    \bigl(f_1(\mathbf{d}), f_2(\mathbf{d}), \ldots, f_{13}(\mathbf{d})\bigr)
\end{align*}

\section{State Transition Function}
\label{app:transition-function}

$P$ can be defined by the following generative process.

\begin{itemize}
    \item If $r < 2$ and $a = k$, for each die $i$:
          \begin{itemize}
              \item if $k_i = 1$, keep $d'_i = d_i$;
              \item else sample $d'_i \sim \mathrm{Unif}\{1,\dots,6\}$ independently.
          \end{itemize}
          Set $c' = c,\ r' = r+1,\ t' = t$.
    \item If $r = 2$ and $a = i$, set $d' = d$, update $c' = \mathrm{score}(c,d,i)$,
          set $r' = 0,\ t' = t+1$.
\end{itemize}

\end{document}